\documentclass[a4paper,10pt,twocolumn]{article}

\usepackage[left=1.5cm,right=1.5cm,top=2cm,bottom=2cm]{geometry}

\usepackage{microtype}
\usepackage{times}

\usepackage{graphicx}
\usepackage[small,bf]{caption} 
\graphicspath{{fig/}}

\usepackage[raggedright]{titlesec}
\titlelabel{\thetitle.\ }
\titleformat*{\section}{\large\bf}  
\titleformat*{\subsection}{\normalsize\bf}  
\titleformat*{\subsubsection}{\normalsize\it}

\usepackage{fancyhdr}
\fancypagestyle{plain}{
    \fancyhf{}
    \fancyfoot[C]{\small\thepage}

    }
\usepackage[cmex10]{amsmath}
\usepackage{amssymb}
\usepackage{cancel}

\renewcommand{\vec}[1]{\boldsymbol{\mathbf{#1}}}
\newcommand{\defeq}{\triangleq}

\usepackage{booktabs}
\usepackage{tabularx}
\usepackage{multirow}
\newcommand{\mytable}{
    \centering
    \small
    \renewcommand{\arraystretch}{1.1}
    }
\newcolumntype{C}{>{\centering\arraybackslash}X}
\newcolumntype{L}{>{\raggedright\arraybackslash}X}

\usepackage{algorithm}  

\usepackage{hyperref}
\hypersetup{colorlinks=true,linktoc=all,citecolor=black,linkcolor=black}

\usepackage{algpseudocode}  
\usepackage{cite}  
\newcommand{\citettt}[2]{{#1}~\cite{#2}}

\title{\huge A segmental framework for fully-unsupervised \\ large-vocabulary speech recognition}
\author{Herman Kamper$^{1}$, Aren Jansen$^2$, Sharon Goldwater$^1$}
\date{\normalsize $^1$School of Informatics, University of Edinburgh and $^2$Google, Inc.\\[2pt]
\texttt{kamperh@gmail.com}, \texttt{arenjansen@google.com}, \texttt{sgwater@inf.ed.ac.uk}}

\usepackage{color}

\newcommand{\edit}[1]{#1}

\usepackage{etoolbox}
\newtoggle{csl}
\togglefalse{csl}

\usepackage{etoolbox}
\makeatletter
\patchcmd{\ttlh@hang}{\parindent\z@}{\parindent\z@\leavevmode}{}{}
\patchcmd{\ttlh@hang}{\noindent}{}{}{}
\makeatother

\begin{document}

\maketitle

\begin{abstract}
    Zero-resource speech technology is a growing research area that aims to develop methods for speech processing in the absence of transcriptions, lexicons, or language modelling text. Early \edit{term discovery} systems focused on identifying isolated recurring \edit{patterns} in a corpus, while more recent full-coverage systems attempt to completely segment and cluster the audio into word-like units---effectively performing unsupervised speech recognition.
\edit{This article presents the first attempt we are aware of to apply such a system to large-vocabulary multi-speaker data. Our}
system uses a Bayesian modelling framework with segmental word representations: each word segment is represented as a fixed-dimensional acoustic embedding obtained by mapping the sequence of feature frames to a single embedding vector. We compare our system on English and Xitsonga datasets to state-of-the-art baselines, using a variety of measures including word error rate (obtained by mapping the unsupervised output to ground truth transcriptions).
\edit{Very high word error rates are reported---in the order of 70--80\% for speaker-dependent and 80--95\% for speaker-independent systems---highlighting the difficulty of this task.}
\edit{Nevertheless, in terms of cluster quality and word segmentation metrics, we}
show that by imposing a consistent top-down segmentation while also using bottom-up knowledge from detected syllable boundaries, both single-speaker and multi-speaker versions of our system outperform a purely bottom-up single-speaker syllable-based approach. We also show that the discovered clusters can be made less speaker- and gender-specific by using an unsupervised autoencoder-like feature extractor to learn better frame-level features (prior to embedding). Our system's discovered clusters are still less pure than those of 
\edit{unsupervised} term discovery systems, but provide far greater coverage.

    \vspace*{4pt}
\end{abstract}

\section{Introduction}

Despite major advances in supervised speech recognition over the last few years, current methods still rely on huge amounts of transcribed speech audio, pronunciation dictionaries, and texts for language modelling.
The collection of these pose a major obstacle for speech technology in under-resourced languages.
In some extreme cases, unlabelled speech data might be the only available resource.
In this \textit{zero-resource} scenario, unsupervised methods are required to learn representations and linguistic structure directly from the speech signal.
Such methods can, for instance, make it possible to search through a corpus of unlabelled speech using voice queries~\cite{park+glass_taslp08}, allow topics within speech utterances to be identified without supervision~\cite{siu+etal_csl14}, or can be used to automatically cluster related spoken documents~\cite{dredze+etal_emnlp10}.
Similar techniques are required to model how human infants acquire language from speech input~\cite{rasanen_speechcom12}, and 
for developing robotic applications that can learn a new language in an unknown environment~\cite{sun+vanhamme_csl13,taniguchi+etal_arxiv15}.

Interest in zero-resource speech processing has grown considerably in the last few years, with two central research areas emerging~\cite{jansen+etal_icassp13,versteegh+etal_interspeech15}.
The first deals with unsupervised representation learning, where the task is to find speech features (often at the frame level) that make it easier to discriminate between meaningful linguistic units (phones or words).
This task has been described as `phonetic discovery', `unsupervised acoustic modelling' and `unsupervised subword modelling', depending on the type of feature representations that are produced.
Approaches include those using bottom-up trained Gaussian mixture models (GMMs) to produce frame-level posteriorgrams~\cite{zhang+glass_icassp10,chen+etal_interspeech15}, using unsupervised hidden Markov models (HMMs) to obtain discrete categorical output in terms of discovered subword units~\cite{varadarajan+etal_acl08,lee+glass_acl12,siu+etal_csl14}, and using unsupervised neural networks~(NNs) to obtain frame-level continuous vector representations~\cite{synnaeve+etal_slt14,renshaw+etal_interspeech15,zeghidour+etal_icassp16}.

The second area of zero-resource research deals with unsupervised segmentation and clustering of speech into meaningful units.
This is important in tasks such as query-by-example search~\cite{zhang+etal_icassp12,levin+etal_icassp15}, where a system needs to find all the utterances in a corpus containing a spoken query, or in unsupervised term discovery (UTD), where a system needs to automatically find repeated word- or phrase-like patterns in a speech collection~\cite{park+glass_taslp08,jansen+vandurme_asru11,lyzinski+etal_interspeech15}.
UTD systems typically find and cluster only isolated acoustic segments, leaving the rest of the data as background.
We are interested in full-coverage segmentation and clustering, where word boundaries and lexical categories are predicted for the entire input.
Several recent studies share this goal~\cite{sun+vanhamme_csl13,chung+etal_icassp13,walter+etal_asru13,lee+etal_tacl15,rasanen+etal_interspeech15}.
Successful full-coverage segmentation systems would perform a type of unsupervised speech recognition. This would allow downstream applications, such as query-by-example search and speech indexing (grouping together related utterances in a corpus), to be developed in a manner similar to when supervised systems are available.
\edit{Unsupervised segmentation and clustering, however, is a daunting task, and current performance lags behind that of even minimally-supervised
systems.
Nevertheless, previous work has shown that high-error rate unsupervised systems can still be used effectively for a wide range of tasks including topic identification and clustering of spoken documents~\cite{gish+etal_interspeech09,dredze+etal_emnlp10,siu+etal_csl14}, speech-to-speech translation of low-resource languages~\cite{martin+etal_asru15,wilkonson+etal_interspeech16}, language recognition~\cite{shum+etal_taslp16}, and in improving purely supervised keyword search systems~\cite{jansen+etal_icassp13}.
}

In previous work~\cite{kamper+etal_taslp16}, we introduced a novel unsupervised segmental Bayesian model for full-coverage segmentation and clustering of small-vocabulary speech.
Other approaches mostly perform frame-by-frame modelling using subword discovery with subsequent or joint word discovery.
In contrast, our approach models whole-word units directly using a fixed-dimensional embedding representation; any potential word segment (of arbitrary length) is mapped to a fixed-length vector, its \textit{acoustic word embedding}, and the model builds a whole-word acoustic model in the embedding space while jointly performing segmentation.
In~\cite{kamper+etal_taslp16} we evaluated the model in an unsupervised digit recognition task using the TIDigits corpus.
Although it was able to accurately segment and cluster the small number of word types (lexical items) in the data, the same system could not be applied directly to multi-speaker data with larger vocabularies.
This was due to the large number of embeddings that had to be computed, and the efficiency of the embedding method itself.

In this paper, we present a new system that uses the same overall
framework as our previous small-vocabulary system, but with several
changes designed to improve efficiency and speaker independence,
allowing us to scale up to large-vocabulary multi-speaker data.
\edit{We believe}
this is the first full-coverage unsupervised speech
recognition system to 
\edit{be applied}
in this regime; previous systems
have either focused on identifying isolated terms~\cite{park+glass_taslp08,jansen+vandurme_asru11,lyzinski+etal_interspeech15}, were
speaker-dependent~\cite{lee+etal_tacl15,rasanen+etal_interspeech15}, or used only a small vocabulary~\cite{walter+etal_asru13,kamper+etal_taslp16}.
\edit{Given this is the first attempt we are aware of, the results reported here will serve as a useful baseline for future work on unsupervised speech recognition of multi-speaker data with realistic vocabularies.}

For our efficiency improvements, we use a bottom-up unsupervised
syllable boundary detection method~\cite{rasanen+etal_interspeech15}
to eliminate unlikely word
boundaries, reducing the number of potential word segments that need
to be considered. We also use a computationally much simpler embedding
approach based on downsampling~\cite{levin+etal_asru13}.

For better speaker-independent performance, we incorporate a
frame-level representation learning method introduced in our previous
work~\cite{kamper+etal_icassp15}: the \textit{correspondence autoencoder} (cAE). The cAE uses noisy
word pairs identified by an unsupervised term detection system to
provide weak supervision for training a deep NN on aligned
frame pairs; features are then extracted from one of the network
layers. In~\cite{kamper+etal_icassp15} we showed that cAE frame-level features outperform
traditional features (MFCCs) and GMM-based representations in a
multi-speaker intrinsic evaluation. Here, we show that the cAE
features also improve performance of our full-coverage multi-speaker
segmentation and clustering system (relative to MFCC features).
\edit{These results are the first to show that unsupervised representation learning can improve a full-coverage zero-resource system.}

We evaluate our approach in both speaker-dependent and
speaker-independent settings on conversational speech datasets from
two languages: English and Xitsonga. Xitsonga is an under-resourced southern
African Bantu language~\cite{devries+etal_speechcom14}.
These datasets were also used as part of the
Zero Resource Speech Challenge (ZRS) at Interspeech 2015~\cite{versteegh+etal_interspeech15} and we
show that our system outperforms competing systems~\cite{versteegh+etal_interspeech15,lyzinski+etal_interspeech15,rasanen+etal_interspeech15} on
several of the ZRS metrics.
\edit{These metrics measure aspects ranging from cluster quality to segmentation performance.}
In particular, we find that by proposing a
consistent segmentation and clustering over a whole utterance, our
approach makes better use of the bottom-up syllabic constraints than
the purely bottom-up syllable-based system of~\cite{rasanen+etal_interspeech15}. Moreover, we
achieve similar $F$-scores for word tokens, types, and boundaries
whether training in a speaker-dependent or speaker-independent mode.

By mapping the unsupervised output to ground truth transcriptions, we also evaluate word error rate (WER), a metric not included in the ZRS Challenge. Our best system has WERs of around 70--80\% for speaker-dependent and 80--95\% for speaker-independent settings. Although these are high error rates, nevertheless our results and analysis should provide useful baselines and guidance for future work in this area.\footnote{Code for this work is available at \url{https://github.com/kamperh/bucktsong_segmentalist}.}


\section{Related work}

Below we first discuss related work on unsupervised representation learning, followed by unsupervised term discovery (which we also compare our approach to), and, finally, full-coverage segmentation and clustering of unlabelled speech.

\subsection{Unsupervised frame-level representation learning}
\label{sec:background_repr}

Unsupervised representation learning, in this context, involves finding a frame-level mapping from input features to a new representation that makes it easier to discriminate between different linguistic units (normally subwords or words).

Early studies used bottom-up approaches operating directly on the acoustics. 
\citettt{Zhang and Glass}{zhang+glass_icassp10} successfully used posteriorgram features from an unsupervised GMM universal background model (UBM) for query-by-example search and term discovery.
Similarly, \citettt{Chen et al.}{chen+etal_interspeech15} used posteriorgrams from a non-parameteric infinite GMM.
Approaches using unsupervised HMMs to perform a bottom-up tokenization of speech include the successive state-splitting algorithm of \citettt{Varadarajan et al.}{varadarajan+etal_acl08}, the more traditional iterative re-estimation and unsupervised decoding procedure of \citettt{Siu et al.}{siu+etal_csl14}, and the non-parameteric Bayesian HMM of \citettt{Lee and Glass}{lee+glass_acl12}.
More recently, NNs have been used for bottom-up representation learning: stacked autoencoders (AEs), a type of unsupervised deep NN that tries to reconstruct its input, has been used in several studies~\cite{zeiler+etal_icassp13,badino+etal_icassp14,badino+etal_interspeech15}.

The above approaches perform representation learning without regard to longer-spanning word- or phrase-like patterns in the data.
In several recent studies, unsupervised term discovery (UTD) is used to automatically discover such patterns; these then serve as weak top-down constraints for subsequent representation learning.
Jansen et al.\ showed that such constraints can be used to train HMMs~\cite{jansen+church_interspeech11} and GMM-UBMs~\cite{jansen+etal_icassp13b} that significantly outperform their pure bottom-up counterparts.
In our own work~\cite{kamper+etal_icassp15}, we proposed the \textit{correspondence autoencoder} (cAE): an AE-like deep NN that incorporates top-down constraints by using aligned frames from discovered words as input-output pairs. 
The model significantly outperformed the top-down GMM-UBM~\cite{jansen+etal_icassp13b} and stacked AEs~\cite{zeiler+etal_icassp13,badino+etal_icassp14} in an intrinsic evaluation: isolated word discrimination.
Since then, several researchers have used such weak top-down supervision in training unsupervised NN-based models~\cite{synnaeve+etal_slt14,thiolliere+etal_interspeech15,zeghidour+etal_icassp16}.
In this paper we show that cAE-learned features also improve performance of our multi-speaker unsupervised segmentation and clustering system.

\subsection{Unsupervised term discovery}
\label{sec:background_utd}

Unsupervised term discovery (UTD) is the task of finding meaningful word- or phrase-like patterns in unlabelled speech data.
Most state-of-the-art UTD systems use a variant of dynamic time warping (DTW), called segmental DTW.
This algorithm, developed by \citettt{Park and Glass}{park+glass_taslp08}, identifies similar sub-sequences within two vector time series, rather than comparing entire sequences as in standard DTW.
In most UTD systems, segmental DTW proposes pairs of matching segments which are then clustered using a graph-based method.
Follow-up work has built on Park and Glass' original method in various ways, for example through improved
feature representations~\cite{zhang+etal_icassp12} or by greatly improving its efficiency~\cite{jansen+vandurme_asru11}.

The baseline provided as part of the lexical discovery track of the Zero Resource Speech Challenge 2015 (ZRS)~\cite{versteegh+etal_interspeech15} is a UTD system based on the earlier work of~\cite{jansen+vandurme_asru11}.
The other UTD submission to the ZRS by \citettt{Lyzinski et al.}{lyzinski+etal_interspeech15} extended the baseline system  using improved graph clustering algorithms.
In our evaluation, we compare to both these systems.
Our approach
shares the property of UTD systems that it has no subword level of representation and operates directly on whole-word representations.
However, instead of representing each segment as a vector time series with variable duration as in UTD, we map each potential word segment to a fixed-dimensional acoustic word embedding; we can then define an acoustic model in the embedding space and use it to compare segments without performing DTW alignment.
Our system also performs full-coverage segmentation and clustering, in contrast to UTD, which segments and clusters only isolated acoustic patterns.

\subsection{Full-coverage segmentation and clustering of speech}
\label{sec:background_full_coverage}

\edit{Early work considered full-coverage word segmentation of transcribed phonemic or phonetic symbol sequences~\cite{goldwater+etal_cognition09,mochihashi+etal_acl09,neubig+etal_interspeech10,heymann+etal_asru13}. This laid the foundation for subsequent efforts to develop methods to entirely segment raw speech into word-like clusters.}
\edit{The approach at the the 2012 JHU CSLP workshop used symbolic word segmentation methods on top of automatically discovered subword units, but this pipelined approach gave very poor performance~\cite{jansen+etal_icassp13}.}
\edit{More recent efforts attempt to segment raw speech directly; }
approaches include using non-negative matrix factorization~\cite{sun+vanhamme_csl13}, using iterative decoding and refinement for jointly training subword HMMs and a lexicon~\cite{chung+etal_icassp13}, and using discrete HMMs to model whole words in terms of discovered subword units~\cite{walter+etal_asru13}.
Below we highlight two studies which have inspired our work in particular.

In~\cite{lee+etal_tacl15}, Lee et al.\ developed a non-parametric hierarchical Bayesian model for full-coverage speech segmentation.
Their model consists of 
a bottom subword acoustic modelling layer, a noisy channel model for capturing pronunciation variability, a syllable layer, and a highest-level word layer.
When applied to speech from single speakers in the MIT Lecture corpus, most words with high TF-IDF scores were successfully discovered.
As in their model, we also follow a Bayesian approach, which is useful for incorporating prior knowledge and for finding sparser solutions~\cite{goldwater+griffiths_acl07}.
However, where~\cite{lee+etal_tacl15} only considered single-speaker data, we additionally evaluate on large-vocabulary multi-speaker data.

Furthermore, in contrast to~\cite{lee+etal_tacl15,chung+etal_icassp13,walter+etal_asru13}, our model operates directly at the whole-word level instead of having both word and subword models.
By taking this different perspective, our segmental whole-word approach 
is a complementary contribution to the field of zero-resource speech processing.
The approach is further motivated by the observation that it 
is often easier to identify cross-speaker similarities between words than between subwords~\cite{jansen+etal_icassp13b}, which is why most UTD systems focus on longer-spanning patterns. There is also evidence that infants are able to segment whole words from continuous speech while still learning phonetic contrasts in their native language~\cite{bortfeld+etal_psychol05,feldman+etal_ccss09}.
A benefit of the segmental embedding approach we use is that segments can be compared directly in a fixed-dimensional embedding space, meaning that word discovery can be performed using standard clustering methods (in our case using a Bayesian GMM acoustic model).
Finally, segmental approaches do not make the frame-level independence assumptions of most of the models above; this assumption has long been argued against~\cite{zweig+nguyen_interspeech10,gillick+etal_asru11}.

The second study we draw from is the ZRS submission of \citettt{R{\"a}s{\"a}nen et al.}{rasanen+etal_interspeech15}, which we use to help scale our approach to larger vocabularies.
Their full-coverage word segmentation system relies on an unsupervised method that predicts boundaries for syllable-like units, and then clusters these units on a per-speaker basis.
Using a bottom-up greedy mapping, reoccurring syllable clusters are then predicted as words.
From here onward we use \textit{syllable} to refer to the syllable-like units detected in the first step of their approach.

In our model, we incorporate the syllable boundary detection method of~\cite{rasanen+etal_interspeech15} (the first component of their system) as a presegmentation method to eliminate unlikely word boundaries.
Both human infants~\cite{emias_jasm99} and adults~\cite{mcqueen_memory98} use syllabic cues for word segmentation, and using such a bottom-up unsupervised syllabifier can therefore be seen as one way to incorporate prior knowledge of the speech signal into a zero-resource system~\cite{versteegh+etal_sltu16}.


\begin{figure*}[t]
    \centering
    \iftoggle{csl}{
        \includegraphics[width=\linewidth]{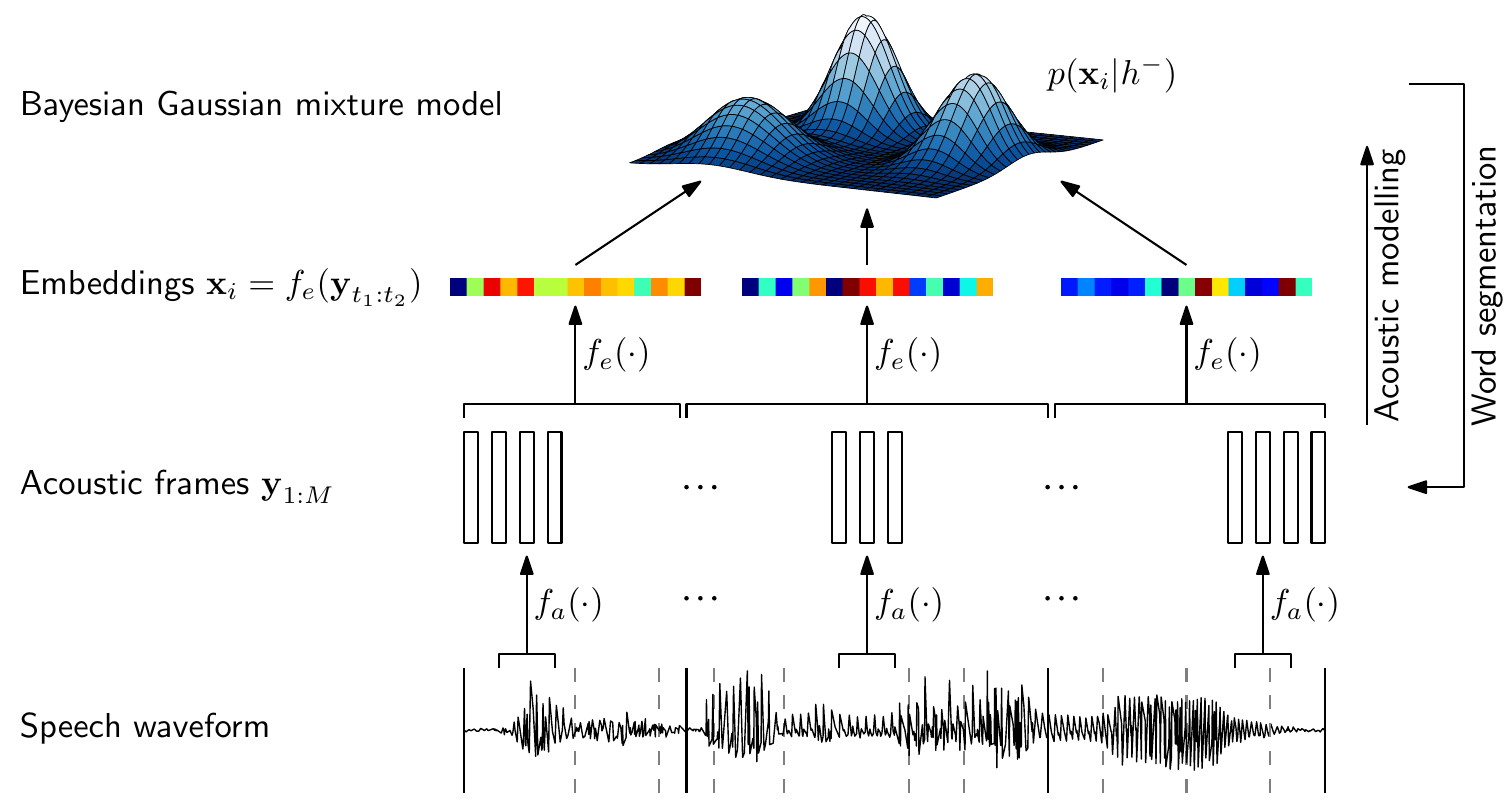}
    }{
        \includegraphics[width=0.675\linewidth]{unsup_repr_wordseg}
    }
    \caption{The large-vocabulary segmental Bayesian model. Dashed lines indicate where word boundaries are allowed according to syllable boundary detection. Function $f_a$ is a frame-level feature extractor, while $f_e$ maps a variable number of frames to a single embedding vector.}
    \label{fig:bucktsong_unsup_repr_wordseg}
\end{figure*}

\section{Large-vocabulary segmental Bayesian model}

In the following we describe our large-vocabulary system in detail, starting with a high-level overview of the model, illustrated in Figure~\ref{fig:bucktsong_unsup_repr_wordseg}.

The model takes as input raw speech (bottom) and converts it to frame-level acoustic features using a sliding window feeding into the feature extracting function $f_a$.
The sequence of frame-level vectors (e.g.\ MFCCs or cAE features) are denoted as $\vec{y}_{1:M} = \vec{y}_1, \vec{y}_2, \ldots, \vec{y}_M$. 
Suppose we have a hypothesis for where word boundaries occur in this stream of features (vertical black lines, bottom of figure).
Each word\footnote{Throughout we use the term \textit{word} to refer to a segment of speech that might in reality correspond to a true word, partial word, phrase or noise, depending on what the system discovers. A more accurate description would be \textit{pseudo term}, but we use \textit{word} instead to match usage in earlier work~\cite{levin+etal_asru13,park+glass_taslp08,kamper+etal_icassp16}.} segment is then mapped to to an \textit{acoustic word embedding} (coloured horizontal vectors in the figure) in a fixed-dimensional space $\mathbb{R}^D$; this is done using the embedding function $f_e$, which takes a sequence of frame-level features as input and outputs a single embedding vector $\vec{x}_i \in \mathbb{R}^D$.
Ideally, embeddings of different instances of the same word type should lie close together in this space.
The different hypothesized word types are then modelled using a whole-word acoustic model: a GMM with Bayesian priors
in the $D$-dimensional embedding space (top of figure).
Effectively, if word boundaries are known, this is simply a clustering model, with every cluster (mixture component) of the GMM corresponding to a discovered word type.

Initially, however, we do not know where words start and end in the
stream of features. But if we have a GMM acoustic model, we can use
this model to segment an utterance by choosing word boundaries that
yield segments (acoustic word embeddings) that have high probability under the
acoustic model. Our full system therefore initializes word boundaries
at random, extracts word embeddings, clusters them using the Bayesian
GMM, and then iteratively re-analyzes each utterance (jointly
re-segmenting it and re-clustering the segments) based on the current
acoustic model.
The result is a complete segmentation of the input speech and a prediction of the component to which every word segment belongs.
The model is implemented as a single blocked Gibbs sampler, and exact details are given next.

\subsection{Segmental Bayesian modelling}
\label{sec:bucktsong_unigram}

\begin{figure}[tbp]
    \centering
    \includegraphics[scale=0.85]{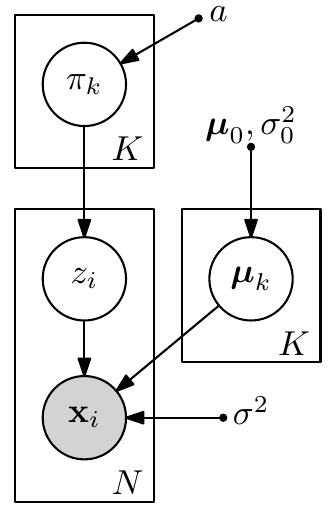}
    \caption{The graphical model of the Bayesian Gaussian mixture model with fixed spherical covariance used as acoustic model.}
    \label{fig:bucktsong_fbgmm}
\end{figure}

Given the embedded word vectors $\mathcal{X} = \{ \vec{x}_i\}_{i = 1}^{N}$ from the current segmentation hypothesis, the acoustic model needs to assign each acoustic word embedding $\vec{x}_i$ to one of $K$ clusters, with each cluster corresponding to a hypothesized word type.
We use a Bayesian GMM as acoustic model, with a conjugate Dirichlet prior over its mixture weights $\vec{\pi}$ and a conjugate diagonal-covariance Gaussian prior over its component means $\left\{ \vec{\mu}_k \right\}_{k = 1}^K$, which allows us to integrate out these parameters.
The model, illustrated in Figure~\ref{fig:bucktsong_fbgmm}, is formally defined as:

\noindent
\begin{minipage}{.45\linewidth}
    \centering
    \begin{alignat}{2}
        &\vec{\pi}  &&\sim \textrm{Dir}\left( a/{K} \vec{1} \right) \label{eq:fbgmm1} \\
        &z_i &&\sim \vec{\pi}  \label{eq:fbgmm2}
    \end{alignat}
    ~
\end{minipage}
\hfill
\begin{minipage}{.45\linewidth}
    \centering
    \begin{alignat}{2}
        &\vec{\mu}_k  &&\sim \mathcal{N} (\vec{\mu}_0, \sigma_0^2 \vec{I})  \label{eq:fbgmm3} \\
        &\vec{x}_i &&\sim \mathcal{N} (\vec{\mu}_{z_i}, \sigma^2 \vec{I})  \label{eq:fbgmm4}
    \end{alignat}
    ~
\end{minipage}

\noindent Latent variable $z_i$ indicates the component to which $\vec{x}_i$ is assigned.
All $K$ components share the same fixed covariance matrix $\sigma^2 \vec{I}$.
The hyperparameters of the mixture components are denoted together as $\vec{\beta} = (\vec{\mu}_0, \sigma_0^2, \sigma^2)$.
\edit{These hyperparameters could potentially be learned themselves, but here we set them by hand based on previous studies, as described in Section~\ref{sec:model_dev_hyperparams}.}

Given $\mathcal{X}$, we infer the component assignments $\vec{z} =
(z_1, z_2, \ldots, z_N)$ using a collapsed Gibbs sampler~\cite{resnik+hardisty_gibbs_tutorial10}.
This is done in turn for each $z_i$ conditioned on all the other current component assignments~\cite{kamper+etal_taslp16}:
\begin{align}
    P(z_i = k|\vec{z}_{\backslash i}, \mathcal{X} ; a, \vec{\beta} ) \propto P(z_i = k|\vec{z}_{\backslash i}; a)  p(\vec{x}_i | \mathcal{X}_{k \backslash i}; \vec{\beta})
    \label{eq:bucktsong_collapsed1}
\end{align}
where $\vec{z}_{\backslash i}$ is all latent component assignments excluding $z_i$
and $\mathcal{X}_{k \backslash i}$ is the set of embedding vectors assigned to component $k$ apart from $\vec{x}_i$.
The first term in~\eqref{eq:bucktsong_collapsed1} can be calculated as:
\begin{equation}
    P(z_i = k|\vec{z}_{\backslash i}; {a}) = \frac{N_{k\backslash i} + a/K}{N + a - 1}
    \label{eq:bucktsong_first_term7}
\end{equation}
where $N_{k \backslash i}$ is the number of embedding vectors from mixture component $k$ without taking $\vec{x}_i$ into account~\cite[p.~843]{murphy}.
This term can be interpreted as a discounted unigram language modelling probability.
The term $p(\vec{x}_i | \mathcal{X}_{k \backslash i}; \vec{\beta})$ in~\eqref{eq:bucktsong_collapsed1} is the posterior predictive of $\vec{x}_i$, which (because of the conjugate prior) is a spherical covariance Gaussian distribution with analytic expressions for its mean and covariance parameters~\cite{murphy_bayesgauss07}\edit{; these expressions are given in~\iftoggle{csl}{\ref{appen:posterior_predictive}}{Appendix~\ref{appen:posterior_predictive}}.}
Intuitively, component assignment sampling in~\eqref{eq:bucktsong_collapsed1} is therefore based on a combination of language model and acoustic scores.

Above we described clustering given the current segmentation. But segmentation and clustering are performed jointly: for the utterance under consideration, a segmentation is sampled using the current acoustic model (marginalizing over cluster assignments for each potential segment), and clusters are then resampled for the newly created segments.
Pseudo-code for the blocked Gibbs sampler that implements this algorithm is given in Algorithm~\ref{alg:bucktsong_gibbs_wordseg}.
The acoustic data is denoted as $\{ \vec{s}_i \}_{i = 1}^S$, where every utterance $\vec{s}_i$ consists of acoustic frames $\vec{y}_{1:M_i}$, and $\mathcal{X}(\vec{s}_i)$ denotes the embedding vectors under the current segmentation for utterance $\vec{s}_i$.
In Algorithm~\ref{alg:bucktsong_gibbs_wordseg}, 
\edit{utterance $\vec{s}_i$ is selected according to a random permutation of all utterances;}
the embeddings from the current segmentation $\mathcal{X}(\vec{s}_i)$ are removed from the Bayesian GMM; a new segmentation is sampled; and finally the embeddings from this new segmentation are added back into the~Bayesian~GMM.
Line~\ref{alg_line:bucktsong_sample_bounds} uses the forward filtering backward sampling dynamic programming algorithm~\cite{scott_jasa02} to sample the new embeddings; details of this step are given in \iftoggle{csl}{\ref{appen:forward_backward}}{Appendix~\ref{appen:forward_backward}}.

\begin{algorithm}[tbp]
\iftoggle{csl}{}{\small}
\begin{algorithmic}[1]
\State Choose an initial segmentation (e.g.\ random).
\For{$j = 1$ to $J$}\Comment{Gibbs sampling iterations}

    \For{$i = $ randperm$(1$ to $S)$} \Comment{Select utterance $\vec{s}_i$}
    
        \State Remove embeddings $\mathcal{X}(\vec{s}_i)$ from acoustic model.\label{alg_line:bucktsong_remove_embeds}
        
        
        \State Resample word boundaries for $\vec{s}_i$, yielding new $\mathcal{X}(\vec{s}_i)$ 
        \label{alg_line:bucktsong_sample_bounds} 

        \For{embedding $\vec{x}_i$ in newly sampled $\mathcal{X}(\vec{s}_i)$} \label{alg_line:bucktsong_component_assignment_start}
        
            \State Sample $z_i$ for embedding $\vec{x}_i$ using~\eqref{eq:bucktsong_collapsed1}.
            \label{alg_line:fbgmm_inside_loop}
        
        \EndFor \label{alg_line:bucktsong_component_assignment_end}

    \EndFor
\EndFor
\end{algorithmic}
\caption{Gibbs sampler of the segmental Bayesian model.}
\label{alg:bucktsong_gibbs_wordseg}
\end{algorithm}

\subsection{Unsupervised syllable boundary detection}

Without any constraints, the input at the bottom of Figure~\ref{fig:bucktsong_unsup_repr_wordseg} could be segmented into any number of possible words using a huge number of possible segmentations.
In~\cite{kamper+etal_taslp16}, potential word segments were therefore required to be between 200~ms and 1~s in duration, and word boundaries were only considered at 20~ms intervals.
This still results in a very large number of possible segments.
Here we instead use a syllable boundary detection method to eliminate unlikely word boundaries, with word candidates spanning a maximum of six syllables.
On the waveform in Figure~\ref{fig:bucktsong_unsup_repr_wordseg}, solid and dashed lines are used to indicate the only positions where boundaries are considered during sampling, as determined by the syllabification method.

\citettt{R{\"a}s{\"a}nen et al.}{rasanen+etal_interspeech15} evaluated several syllable boundary detection algorithms, and we use the best of these.
First the envelope of the raw waveform is calculated by downsampling the rectified signal and applying a low-pass filter.
Inspired by neuropsychological studies which found that neural oscillations in the auditory cortex occur at frequencies similar to that of the syllabic rhythm in speech, the calculated envelope is used to drive a discrete time oscillation system with a centre frequency of typical syllabic rhythm. 
\edit{This discrete time system is used to mathematically model the damped harmonic oscillations in the auditory system, which is hypothesized to match syllabic rhythm.}
Minima in the oscillator's amplitude give the predicted syllable boundaries. 
In this work, we use the syllabification code kindly provided by the authors of~\cite{rasanen+etal_interspeech15} without any modification and with the default parameter settings.


\subsection{Acoustic word embeddings and unsupervised representation learning}
\label{sec:downsampling}

A simple and fast approach to obtain acoustic word embeddings is to uniformly {downsample} so that any segment is represented by the same fixed number of vectors~\cite{ossama+etal_interspeech13,levin+etal_asru13}.
A similar approach is to divide a segment into a fixed number of intervals and average the frames in each interval~\cite{lee+lee_taslp13,rasanen+etal_interspeech15}.
The downsampled or averaged frames are then flattened to obtain a single fixed-length vector.
Although these very simple approaches are less accurate at word discrimination than the approach used before in~\cite{kamper+etal_taslp16}, they 
have been effectively used in several studies, including~\cite{rasanen+etal_interspeech15}, and are computationally much more efficient.
Here we use \textit{downsampling} as our acoustic word embedding function $f_e$ in Figure~\ref{fig:bucktsong_unsup_repr_wordseg}; we keep ten equally-spaced vectors from a segment, and use a Fourier-based method for smoothing \edit{to deal with cases where segments are not exactly divisible}~\cite{levin+etal_asru13}.

Figure~\ref{fig:bucktsong_unsup_repr_wordseg} shows that 
$f_e$ takes as input a sequence of frame-level features from the feature extracting function $f_a$. 
One option for $f_a$ is to simply use MFCCs.
As an alternative, we incorporate unsupervised representation learning (Section~\ref{sec:background_repr}) into our approach by using the cAE as a feature extractor.
Complete details of the cAE are given in~\cite{kamper+etal_icassp15}, but we briefly outline the training procedure here.
The UTD system of~\cite{jansen+vandurme_asru11} is used to discover word pairs which serve as weak top-down supervision.
The cAE operates at the frame level, so 
the word-level constraints are converted to frame-level constraints
by aligning 
each word pair using DTW. Taken together across all discovered pairs, this results in a set of $F$ frame-level pairs $\left\{ \left(\vec{y}_{i,a}, \vec{y}_{i,b} \right) \right\}_{i=1}^{F}$.
Here, each frame is a single MFCC vector.
For every pair $\left(\vec{y}_{a}, \vec{y}_{b}\right)$, $\vec{y}_{a}$ is presented as input to the cAE while $\vec{y}_{b}$ is taken as output, and vice versa. 
The cAE consists of several non-linear layers which are initialized by pretraining the network as a standard autoencoder. 
The cAE is then tasked with reconstructing $\vec{y}_{b}$ from $\vec{y}_{a}$, using the loss $\left|\left|\vec{y}_{b} - \vec{y}_{a}\right|\right|^2$. 
To use the trained network as a feature extractor $f_a$, the activations in one of its middle layers are taken as the new feature representation. 

\section{Experiments}

\subsection{Experimental setup}
\label{sec:bucktsong_exp_setup}

We use three datasets, summarized in Table~\ref{tbl:bucktsong_data}. 
The first two are disjoint subsets extracted from the Buckeye corpus of conversational English~\cite{pitt+etal_speechcom05}, while the third is a portion of the Xitsonga section of the NCHLT corpus of languages spoken in South Africa~\cite{devries+etal_speechcom14}.
Xitsonga is a Bantu language spoken in southern Africa; although it is considered under-resourced, more than five million people use it as their first language.\footnote{\url{http://www.ethnologue.com/language/tso}} 

\iftoggle{csl}{\begin{table}[tbp]}{\begin{table}[b]}
    \mytable
    \caption{Statistics for the datasets used here. Sets have an equal number of female and male speakers. The last column is an average.}
    \begin{tabularx}{\linewidth}{@{}lCCCCC@{}}
        \toprule
        Dataset & Duration (hours) & No. of speakers & Word tokens & Word types & Types per spk. \\
        \midrule
        English1 & 6.0 & 12 & 89\,681 & 5129 & 1104 \\
        English2 & 5.0 & 12 & 69\,543 & 4538 & 966 \\
        Xitsonga & 2.5 & 24 & 19\,848 & 2288 & 333 \\
        \bottomrule
    \end{tabularx}
    \label{tbl:bucktsong_data}
\end{table}

The two sets extracted from Buckeye, referred to as English1 and English2, respectively contain six and five hours of speech, each from twelve speakers (six female and six male).
The Xitsonga dataset consists of 2.5 hours of speech from 24 speakers (twelve female, twelve male).
English2 and the Xitsonga data were used as test sets in the ZRS challenge, so we can compare our system to others using the same data and evaluation framework~\cite{versteegh+etal_interspeech15}.
English1 was extracted for development purposes from a disjoint portion of Buckeye to match the distribution of speakers in English2.
For all three sets, speech activity regions are taken from forced alignments of the data, as was done in the ZRS.
From Table~\ref{tbl:bucktsong_data}, the average duration of a word in an English set is around 250~ms, while for Xitsonga it is about 450~ms.

Our model is unsupervised, which means that the concepts of training and test data become blurred. We run our model on
all sets separately---in each case, unsupervised modelling and evaluation is performed on the same set.
English1 is the only set used for any development (specifically for setting hyperparameters) in any of the experiments; both English2 and Xitsonga are treated as unseen final test sets.
This allows us to see how hyperparameters generalize within language on data of similar size, as well as across language on a corpus with very different characteristics.

\subsection{Evaluation}
\label{sec:metrics}

The evaluation of zero-resource systems that segment and cluster speech is a research problem in itself~\cite{ludusan+etal_lrec14}.
We use a range of metrics that have been proposed before, all performing some mapping from the discovered structures to ground truth forced alignments of the data, as illustrated in Figure~\ref{fig:bucktsong_evaluation}.

\textbf{Average cluster purity} first aligns every discovered {token} to the ground truth word token with which it overlaps most. In Figure~\ref{fig:bucktsong_evaluation} the token assigned to cluster 931 would be mapped to the true word `yeah', and the 477-token mapped to `mean'.
Every discovered word {type} (cluster) is then mapped to the most common ground truth word type in that cluster.
E.g.\ if most of the other tokens in cluster 931 are also labelled as `yeah', then cluster 931 would be labelled as `yeah'.
Average purity is then defined as the total proportion of correctly mapped tokens in all clusters.
For this metric, more than one cluster may be mapped to a single ground truth type (i.e.\ many-to-one)~\cite{sun+vanhamme_csl13}.

\iftoggle{csl}{\begin{figure}[tbp]}{\begin{figure}[b]}
    \centering
    \iftoggle{csl}{\includegraphics[width=0.8\linewidth]{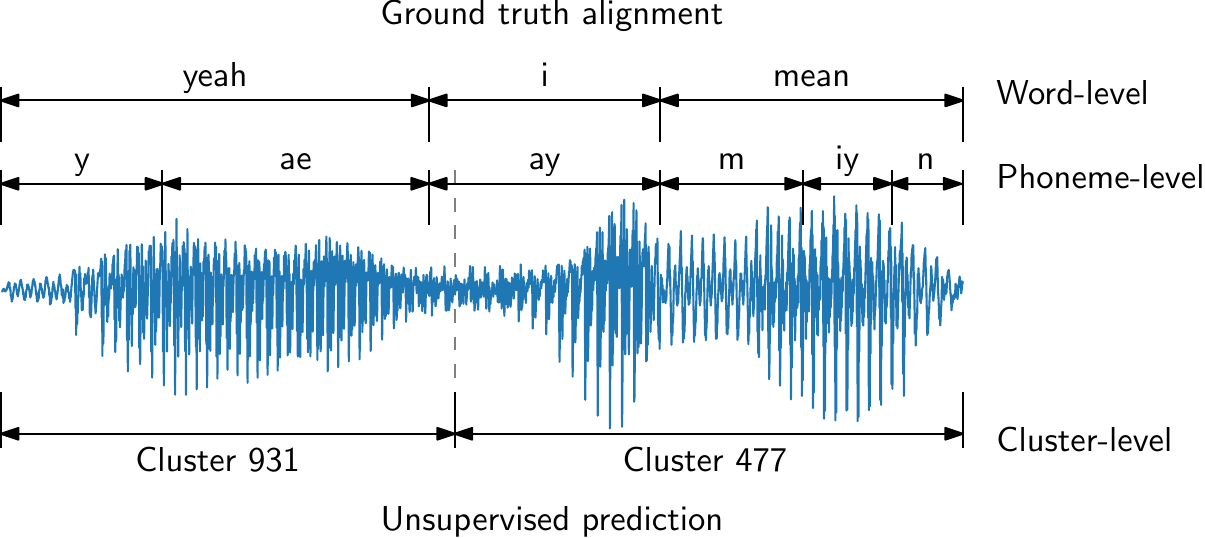}}{\includegraphics[width=\linewidth]{evaluation}}
    \caption{Illustration of the mapping of clusters to true labels for evaluation. Ground truth alignments are shown at the top, with actual output from speaker-dependent BayesSegMinDur-cAE at the bottom.}
    \label{fig:bucktsong_evaluation}
\end{figure}

\textbf{Unsupervised word error rate (WER/WER$_\text{m}$)} uses a similar word-level mapping and then aligns the mapped decoded output from
a system to the ground truth transcriptions~\cite{chung+etal_icassp13,walter+etal_asru13}.
Based on this alignment we calculate $\textrm{WER} = \frac{S + D + I}{N}$, with $S$ the number of substitutions, $D$ deletions, $I$ insertions, and $N$ the tokens in the ground truth.
The cluster mapping can be done in one of two ways: many-to-one, where more than one cluster can be assigned the same word label (as in purity), or using a greedy one-to-one mapping, where at most one cluster is mapped to a ground truth word type.
The latter, which we denote simply as WER, might leave some cluster unassigned and these are counted as errors~\cite{kamper+etal_taslp16}.
For the former, denoted as WER$_\text{m}$, all clusters are labelled.
Depending on the downstream speech task, it might be acceptable to have multiple clusters that correspond to the same true word; WER penalizes such clusters, while WER$_\text{m}$ does not.
WER is a useful metric since it is easily interpretable and well-known in the speech community.

\textbf{Normalized edit distance (NED)} is the first of the ZRS metrics (the rest follow).
These metrics use a phoneme-level mapping: each discovered token is mapped to the sequence of ground truth phonemes of which at least 50\% or 30~ms are covered by the discovered segment\edit{, i.e.\ if a phoneme overlaps with either 30~ms or 50\% of its duration with the discovered segment, it becomes part of the phoneme sequence to which that segment is mapped}~\cite{ludusan+etal_lrec14,versteegh+etal_interspeech15}.
In Figure~\ref{fig:bucktsong_evaluation}, the 931-token would be mapped to \mbox{/y ae/} and the 477-token to \mbox{/ay m iy n/}.
For a pair of discovered segments, the edit distance between the two phoneme strings is divided by the maximum of the
length of the two strings. This is averaged over all pairs predicted to be of the same type (cluster), to obtain the final NED score. If all segments in each cluster have the same phoneme string, then $\textrm{NED} = 0$, while if all phonemes are different, $\textrm{NED} = 1$.
NED is useful in that it does not make the assumptions that the discovered segments need to correspond to true words (as in cluster purity and WER), and it only considers the patterns returned by a system (so it does not require full coverage, as WER does).
As an example, if a cluster contains \mbox{/m iy/} from a realization of the word `meaningful' and a token \mbox{/m iy n/} from the true word `mean', then NED would be $1/3$ for this two-token cluster.

\textbf{Word boundary precision, recall, $F$-score} are calculated by comparing word boundary positions proposed by a system to those from forced alignments of the data, falling within some tolerance.
A tolerance of 20~ms is mostly used~\cite{lee+etal_tacl15}, but for the ZRS the tolerance is 30 ms or 50\% of a phoneme (to match the mapping).
In Figure~\ref{fig:bucktsong_evaluation} the detected boundary (dashed line) would be considered correct if it is within
the tolerance from the true word boundary between `yeah' and `i'.

\textbf{Word token precision, recall, $F$-score} compare how accurately proposed word tokens match ground truth word tokens in  the data.
In contrast to the word boundary scores, both boundaries of a predicted word token need to be correct.
In Figure~\ref{fig:bucktsong_evaluation}, the system would receive credit for the 931-token since it is mapped to \mbox{/y ae/} and therefore match the ground truth word token `yeah'.  However, the system would be penalized for the 477-token (mapped to \mbox{/ay m iy n/}) since it fails to predict word tokens corresponding to \mbox{/ay/} and \mbox{/m iy n/} (the ground truth words `i' and `mean').
Both the word boundary and word token metrics give a measure of how accurately a system is segmenting its input into word-like units.

\textbf{Word type precision, recall, $F$-score} compare the set of distinct phoneme mappings from the tokens returned by a system to the set of true word types in the ground truth alignments. If any discovered word token maps to a phoneme sequence that is also found as a word in the ground truth vocabulary, the system is credited for a correct discovery of that word type. 
For example if the type \mbox{/y ae/} (as in `yeah') occurs in the ground truth alignment, the system needs to return at least one token that is mapped to \mbox{/y ae/}.

We evaluate our model in 
both speaker-dependent and speaker-independent settings.
Multiple speakers make it more difficult to discover accurate clusters: non-matching linguistic units might be more similar within-speaker than matching units across speakers.
For the speaker-dependent case, the model is run and scores are computed on each speaker individually, then performance is averaged over speakers.
In the speaker-independent case, the system is run and scores computed over the entire multi-speaker dataset at once.
This typically results in worse purity, NED and WER$_\text{m}$ scores since the task is more difficult and clusters are noisier.
WER is affected even more severely due to the one-to-one mapping that it uses; if there are two perfectly pure clusters that contain tokens from the same true word, but the two clusters are also perfectly speaker-dependent, then only one of these clusters would be mapped to the true word type and the other would be counted as errors.
Despite the adverse effect on these metrics, it is of practical importance to evaluate a zero-resource system in the speaker-independent setting.

\subsection{Model development and hyperparameters}
\label{sec:model_dev_hyperparams}

Most model hyperparameters are set according to previous work \edit{(as referenced below)}.
Any changes are based exclusively on performance on English1. 

Training parameters for the cAE (Section~\ref{sec:downsampling}) are based on~\cite{kamper+etal_icassp15,renshaw+etal_interspeech15}.
The model is pretrained \edit{as a standard autoencoder} on all data (in a particular set) for 5 epochs using minibatch stochastic gradient descent with a batch size of 2048 and a fixed learning rate of $2\cdot10^{-3}$.
Subsequent correspondence training is performed for 120 epochs using a learning rate of $32\cdot10^{-3}$.
Each pair is presented in both directions as input and output.
Pairs are extracted using the UTD system of~\cite{jansen+vandurme_asru11}: for English1, 14\,494 word pairs are discovered; for English2, 10\,769 pairs; and for Xitsonga, 6979.
The cAE is trained on each of these sets separately.
In all cases, the model consists of nine hidden layers of 100 units each, except for the eighth layer which is a bottleneck layer of 13 units.
We use $\tanh$ as non-linearity.
The position of the bottleneck layer is based on intrinsic evaluation on English1.
Although it is common in NN speech systems to use nine or eleven sliding frames as input, we use 
single-frame~\edit{cepstral mean and variance normalized} MFCCs with first and second order derivatives (39-dimensional), as also done in~\cite{kamper+etal_icassp15,renshaw+etal_interspeech15}.
For feature extraction, the cAE is cut at the bottleneck layer, resulting in 13-dimensional output (chosen to match the dimensionality of the static MFCCs). 
For both the MFCC and cAE acoustic word embeddings, we downsample a segment to ten frames, resulting in 130-dimensional embeddings. 
As in~\cite{kamper+etal_slt14,kamper+etal_interspeech15,kamper+etal_taslp16}, embeddings are normalized to \edit{have unit length}. 

\iftoggle{csl}{
    \begin{table*}[b]
}{
    \begin{table*}[t]
}
    \mytable
    \caption{Performance on the three datasets for speaker-dependent models.}
    \iftoggle{csl}{
        \begin{tabularx}{\linewidth}{@{}l@{\ }c@{\ }CCCCCC@{}}
    }{
        \begin{tabularx}{0.9\linewidth}{@{}lcCCCCCC@{}}
    }
        \toprule
        & & \multicolumn{3}{c}{One-to-one WER (\%)} & \multicolumn{3}{c}{Many-to-one WER$_\text{m}$ (\%)} \\
        \cmidrule(l){3-5} \cmidrule(l){6-8}
        Model & Embeds. & English1 & English2 & Xitsonga & English1 & English2 & Xitsonga \\
        \midrule
        SyllableBayesClust & MFCC & 93.3 & 94.1 & 140.3 & 72.4 & 76.1 & 134.5 \\
        BayesSeg & MFCC & 89.2 & 88.8 & 116.2 & 68.3 & 70.5 & 109.5 \\
        BayesSegMinDur & MFCC & 83.7 & 82.8 & 78.9 & 67.6 & 68.3 & 71.7 \\
        BayesSeg & cAE & 89.3 & 89.3 & 107.9 & 70.0 & 73.0 & 100.5 \\
        BayesSegMinDur & cAE & 85.2 & 84.1 & 75.9 & 70.6 & 71.2 & 68.8 \\
        \bottomrule
    \end{tabularx}    
    \label{tbl:bucktsong_sd_wer}
\end{table*}

For the acoustic model (Section~\ref{sec:bucktsong_unigram})
we use the following hyperparameters, as in~\cite{kamper+etal_slt14,kamper+etal_interspeech15,kamper+etal_taslp16}: all-zero vector for $\vec{\mu}_0$, $\sigma_0^2 = \sigma^2/\kappa_0$, $\kappa_0 = 0.05$ and $a = 1$.
For MFCC embeddings we use $\sigma^2 = 1\cdot10^{-3}$ for the fixed shared spherical covariance matrix, while for cAE embeddings we use $\sigma^2 = 1\cdot10^{-4}$. This was based on speaker-dependent English1 performance.
We found that $\sigma^2$ is one of the parameters most sensitive to the input representation and often requires tuning; generally, however, it is robust if it is chosen small enough (in the ranges used here).

We use the oscillator-based syllabification system of \citettt{R{\"a}s{\"a}nen et al.}{rasanen+etal_interspeech15} without modification.
Word candidates are limited to span a maximum of six syllables.
One difficulty is to decide beforehand how many potential word clusters (the number of components $K$ in the acoustic model) we need.
Here we follow the same approach as in~\cite{rasanen+etal_interspeech15}: we choose $K$ as a proportion of the number of discovered syllable tokens.
For the speaker-dependent settings, we set $K$ as $20\%$ of the number of syllables, based on English1 performance.
On average, this amounts to $K = 1549$ on English1, $K = 1195$ on English2, and $K = 298$ on Xitsonga.
Compared to the average number of word types per speaker shown in Table~\ref{tbl:bucktsong_data}, these numbers are higher for the English sets and slightly lower for Xitsonga.
For speaker-independent models, we use $5\%$ of the syllable tokens, amounting to $K = 4647$ on English1, $K = 3584$ on English2, and $K = 1789$ on Xitsonga.
These are lower than the true number of total word types shown in Table~\ref{tbl:bucktsong_data}.
On English1, speaker-independent performance did not improve when using a larger $K$ and inference was much slower.

To improve sampler convergence, we use simulated annealing~\cite{kamper+etal_taslp16}.
We found that convergence is improved by first running the sampler in Algorithm~\ref{alg:bucktsong_gibbs_wordseg} without sampling boundaries.
In all experiments we do this for 15 iterations.
Subsequently, the complete sampler is run for $J = 15$ Gibbs sampling iterations with 3 annealing steps. 
Word boundaries are initialized randomly by setting boundaries at allowed locations with a $0.25$ probability.

Given the common setup above, we consider three variants of our approach:

\textbf{BayesSeg} is the most general segmental Bayesian model. In this model, a word segment can be of any duration, as long as it spans less than six syllables.

\textbf{BayesSegMinDur} is the same as BayesSeg, but requires word candidates to be at least 250~ms in duration; on English1, this improved performance on several metrics. Such a minimum duration constraint is also used in most UTD systems~\cite{park+glass_taslp08,jansen+vandurme_asru11}.

\textbf{SyllableBayesClust} clusters the discovered syllable tokens using the Bayesian GMM, but does not sample word boundaries. It can be seen as a baseline for the two models above, where segmentation is turned off and the detected syllable boundaries are set as initial (and permanent) word boundaries.
All word candidates therefore span a single syllable in this model.

\subsection{Results: Word error rates and analysis}
\label{sec:bucktsong_results_own}

\subsubsection*{Speaker-dependent models}

Table~\ref{tbl:bucktsong_sd_wer} shows one-to-one and many-to-one WERs for the different speaker-dependent models on the three datasets.
The trends in WER using one-to-one and many-to-one mappings are similar, with the absolute performance of the latter consistently better by around 10\% to 20\% absolute.
The performance on Xitsonga varies much more dramatically than on the English datasets, with WER ranging from around 140\% to 75\% and WER$_\text{m}$ from 135\% to 69\%.\footnote{From its definition, WER is more than 100\% if there are more substitutions, deletions and insertions than ground truth tokens.}
Table~\ref{tbl:bucktsong_data} shows that the characteristics of the Xitsonga data are quite different from the English sets.
For the speaker-dependent case here, much less data is available per Xitsonga speaker (just over six minutes on average) than for an English speaker (more than ten minutes), which might (at least partially) explain why error rates vary much more dramatically on Xitsonga.
Moreover, there is a much higher proportion of multisyllabic words in Xitsonga~\cite{rasanen+etal_interspeech15}, as reflected in the average duration of words which is almost twice as long in the Xitsonga than in the English data (Section~\ref{sec:bucktsong_exp_setup}).

Comparing the results for the three
systems using MFCC features indicates that, on all three datasets,
allowing the system to infer word boundaries across multiple syllables
(BayesSeg) yields better performance than treating each syllable as a
word candidate (SyllableBayesClust). Incorporating a minimum duration
constraint (BayesSegMinDur) improves performance further.
The relative
differences between these systems are much more pronounced in
Xitsonga, presumably due to the higher proportion of multisyllabic
words.
\edit{Despite the high error rates, this analysis nevertheless shows the benefits of top-down segmentation and minimum duration constraints; using bootstrap confidence interval estimation~\cite{bisani+ney_icassp04}\footnote{Sampling with replacement at the utterance level, $B = 1000$ bootstrap samples of a dataset are generated. For a single system, WER can then be calculated for each of these samples in order to estimate the spread of the WER around its mean. To compare two systems, the difference in WER is calculated when evaluating both systems on each of the samples, giving an estimate of the probability of improvement of one system over another. See~\cite{bisani+ney_icassp04} for complete details.}, these improvements of BayesSeg over SyllableBayesClust and of BayesSegMinDur over BayesSeg were found to be statistically significant at the 99.9\% level for all three datasets and for both the WER and WER$_\text{m}$ metrics.}

Table~\ref{tbl:bucktsong_sd_wer} also shows that in most cases the cAE features perform
similarly to MFCC features in these speaker-dependent systems,
although there is a large improvement in Xitsonga for the BayesSeg
system when switching to cAE features (from 116.2\% to 107.9\% in WER and from 109.5\% to 100.5\% in WER$_\text{m}$\edit{, again significant at the 99.9\% level}).

To get a better insight into the types of errors that the models make, Tables~\ref{tbl:bucktsong_sd_english2} and~\ref{tbl:bucktsong_sd_tsonga} give a breakdown of word boundary detection scores, individual error rates, and average cluster purity on English2 and Xitsonga, respectively.
\edit{Bootstrap estimates of two standard deviations around each WER are also given, indicating the range in which the true WER lie with 95\% probability~\cite{bisani+ney_icassp04}.}
A word boundary tolerance of 20~ms is used~\cite{lee+etal_tacl15}, with a greedy one-to-one mapping for calculating error rates.
SyllableBayesClust gives an upper-bound for word boundary recall since every syllable boundary is set as a word boundary.
The low recall (28.9\% and 24.8\%)
could potentially be improved by using a better syllabification method, but we leave such an investigation for future work.

\iftoggle{csl}{
    \begin{table*}[t]
}{
    \begin{table*}[t]
}
    \mytable
    \caption{A breakdown of the errors on English2 for the speaker-dependent models in Table~\ref{tbl:bucktsong_sd_wer}. The word boundary detection tolerance is 20~ms. The greedy one-to-one cluster mapping is used for error rate computations\edit{, and bootstrap estimates of two standard deviations (95\% of values) around each WER are shown.}}
    \iftoggle{csl}{
        \begin{minipage}[c]{1.15\linewidth}
        \begin{tabularx}{\linewidth}{@{}l@{\ }c@{\ }CCCCCC@{\ \ }cc@{}}
        \toprule
        & & \multicolumn{3}{c}{Word bound.\ (\%)} & \multicolumn{4}{c}{Errors (\%)} & Purity \\
    }{
        \begin{tabularx}{1.0\linewidth}{@{}lcCCCCCCcc@{}}
        \toprule
        & & \multicolumn{3}{c}{Word boundary (\%)} & \multicolumn{4}{c}{Errors (\%)} & Purity \\
    }
        \cmidrule(lr){3-5} \cmidrule(l){6-9} \cmidrule(l){10-10}
        Model & Embeds. & Prec. & Rec. & $F$  & Sub. & Del. & Ins. & WER & Avg.~(\%) \\
        \midrule
        SyllableBayesClust & MFCC & 27.7 & 28.9 & 28.3 & 63.8 & 13.6 & 16.7 & $94.1\pm0.4$ & 42.0 \\
        BayesSeg & MFCC & 29.3 & 26.3 & 27.7 & 59.3 & 18.3 & 11.2 & $88.8\pm0.4$ & 45.1\\
        BayesSegMinDur & MFCC & 31.5 & 12.4 & 17.8 & 38.3 & 43.2 & 1.3 & $82.8\pm0.3$ & 56.0 \\
        BayesSeg & cAE & 29.1 & 22.8 & 25.6 & 55.7 & 24.3 & 9.3 & $89.3\pm0.3$ & 43.9 \\
        BayesSegMinDur & cAE & 30.9 & 10.0 & 15.1 & 35.4 & 47.7 & 1.0 & $84.1\pm0.3$ & 55.5 \\
        \bottomrule
    \end{tabularx}
    \label{tbl:bucktsong_sd_english2}
    \iftoggle{csl}{\end{minipage}}{}
\end{table*}


\begin{table*}[t]
    \mytable
    \caption{A breakdown of the errors on Xitsonga for the speaker-dependent models in Table~\ref{tbl:bucktsong_sd_wer}. The word boundary detection tolerance is 20~ms. The greedy one-to-one cluster mapping is used for error rate computations\edit{, and bootstrap estimates of two standard deviations (95\% of values) around each WER are shown.}}
    \iftoggle{csl}{
        \begin{minipage}[c]{1.15\linewidth}
        \begin{tabularx}{\linewidth}{@{}l@{\ }c@{\ }CCCCCC@{\ \ }cc@{}}
        \toprule
        & & \multicolumn{3}{c}{Word bound.\ (\%)} & \multicolumn{4}{c}{Errors (\%)} & Purity \\
    }{
        \begin{tabularx}{1.0\linewidth}{@{}lcCCCCCCcc@{}}
        \toprule
        & & \multicolumn{3}{c}{Word boundary (\%)} & \multicolumn{4}{c}{Errors (\%)} & Purity \\
    }
        \cmidrule(lr){3-5} \cmidrule(l){6-9} \cmidrule(l){10-10}
        Model & Embeds. & Prec. & Rec. & $F$  & Sub. & Del. & Ins. & WER & Avg.~(\%) \\
        \midrule
        SyllableBayesClust & MFCC & 12.4 & 24.8 & 16.5 & 55.8 & 2.1 & 82.4 & $140.3\pm1.5$ & 33.1 \\
        BayesSeg & MFCC & 12.4 & 20.3 & 15.4 & 53.5 & 6.0 & 56.6 & $116.2\pm1.3$ & 36.8 \\
        BayesSegMinDur & MFCC & 11.8 & 10.8 & 11.3 & 43.2 & 21.2 & 14.5 & $78.9\pm0.7$ & 50.1 \\
        BayesSeg & cAE & 12.4 & 18.3 & 14.8 & 50.2 & 9.7 & 47.9 & $107.9\pm1.2$ & 40.0\\
        BayesSegMinDur & cAE & 11.5 & 8.9 & 10.0 & 38.3 & 27.9 & 9.7 & $75.9\pm0.7$ & 63.7\\
        \bottomrule
    \end{tabularx}
    \label{tbl:bucktsong_sd_tsonga}
    \iftoggle{csl}{\end{minipage}}{}
\end{table*}

Table~\ref{tbl:bucktsong_sd_english2} shows that on English2, the MFCC-based BayesSeg and BayesSegMinDur models under-segment compared to SyllableBayesClust, causing systematically poorer word boundary recall and $F$-scores and an increase in deletion errors.
However, this is accompanied by large reductions in substitution and insertion error rates, resulting in overall WER improvements and 
more accurate clusters when boundaries are inferred (45.1\% purity, BayesSeg-MFCC) rather than using fixed syllable boundaries (42\%, SyllableBayesClust), with further improvements when not allowing short word candidates (56\%, BayesSegMinDur-MFCC).

In contrast to English2, Table~\ref{tbl:bucktsong_sd_tsonga} shows that on Xitsonga, SyllableBayesClust heavily over-segments causing a large number of insertion errors.
This is not surprising since every syllable is treated as a word, while most of the true Xitsonga words are multisyllabic.
At the cost of more deletions and poorer word boundary detection, BayesSeg-MFCC and BayesSegMinDur-MFCC systematically reduces substitution and insertion errors, again resulting in better overall WER and average cluster purity.
Where the cAE-based
models on English2 performed more-or-less on par with their MFCC counterparts, on Xitsonga the cAE embeddings yield large improvements on some metrics: by switching to cAE embeddings, the WER of BayesSeg improves by 8.3\% absolute, while average cluster purity is 13.6\% better for BayesSegMinDur.

\subsubsection*{Speaker-independent models}

Table~\ref{tbl:bucktsong_si_wer} gives the performance of different speaker-independent models.
Compared to the speaker-dependent results of Table~\ref{tbl:bucktsong_sd_wer}, performance is worse for all models and datasets.
\edit{Dealing with multiple speakers is clearly challenging for these unsupervised systems. Nevertheless, the analysis still allows us to compare the different variants of our approach.}
As in the speaker-dependent case, BayesSegMinDur is the best performing MFCC system, followed by BayesSeg, and SyllableBayesClust performs worst\edit{; again these differences are significant at the 99.9\% level}.
In the speaker-dependent experiments, some MFCC-based models slightly outperformed their cAE counterparts.
Here, however, the WERs of cAE models are identical or improved in all cases;
for Xitsonga in particular, improvements are obtained by using cAE features in both BayesSeg (improvement of 26.3\% absolute in WER) and BayesSegMinDur (7.4\%).
The cAE-based BayesSegMinDur model is the only speaker-independent Xitsonga model with a WER less than 100\%.
Again, by allowing more than one cluster to be mapped the same true
word type, WER$_\text{m}$ scores are lower than WER. On English, the cAE-based
models do not yield better WER$_\text{m}$ than their MFCC counterparts, probably
because WER$_\text{m}$ does not penalize for creating separate speaker- or
gender-specific clusters (these would just get mapped to the same
word for scoring). Nevertheless, the cAE features still yield large
improvements in Xitsonga. Word boundary scores and substitution,
deletion and insertion errors (not shown) follow a similar pattern to
that of the speaker-dependent models.
\edit{Bootstrap estimates of the spread around the individual WERs were in the same order as those in Tables~\ref{tbl:bucktsong_sd_english2} and~\ref{tbl:bucktsong_sd_tsonga}; the SyllableBayesClust Xitsonga system has the biggest spread with the true WER lying in $167.2\pm1.6\%$ with 95\% probability.}

To better illustrate the benefits of 
unsupervised representation learning, Table~\ref{tbl:bucktsong_si_purity} shows general purity measures 
for the speaker-independent MFCC- and cAE-based BayesSegMinDur models.
Average cluster purity is as defined before.
Average speaker purity is similarly defined, but instead of considering the mapped ground truth label of a segmented token, it considers the speaker who produced it: speaker purity is 100\% if every cluster contains tokens from a single speaker, while it is $1/12 = 8.3\%$ if all clusters are completely speaker balanced 
for the English sets and $1/24 = 4.2\%$ for Xitsonga.
Average gender purity is similarly defined: it is 100\% if every cluster contains tokens from a single gender, while $1/2 = 50\%$ indicates a perfectly gender-balanced cluster.
Ideally, a speaker-independent system should have high cluster purity and low speaker and gender purities.
Table~\ref{tbl:bucktsong_si_purity} indicates that for all three datasets, cAE-based embeddings are less speaker and gender discriminative, and have higher or similar cluster purity compared to the MFCC-based embeddings.


\begin{table*}[tbp]
    \mytable
    \caption{Performance on the three datasets for speaker-independent models.}
    \iftoggle{csl}{
        \begin{tabularx}{\linewidth}{@{}l@{\ }c@{\ }CCCCCC@{}}
    }{
        \begin{tabularx}{0.8\linewidth}{@{}lcCCCCCC@{}}
    }
        \toprule
        & & \multicolumn{3}{c}{One-to-one WER (\%)} & \multicolumn{3}{c}{Many-to-one WER$_\text{m}$ (\%)} \\
        \cmidrule(l){3-5} \cmidrule(l){6-8}
        Model & Embeds. & English1 & English2 & Xitsonga & English1 & English2 & Xitsonga \\
        \midrule
        SyllableBayesClust & MFCC & 105.1 & 106.5 & 167.2 & 86.4 & 89.6 & 149.2 \\
        BayesSeg & MFCC & 101.7 & 102.1 & 148.3 & 83.4 & 85.6 & 131.3 \\
        BayesSegMinDur & MFCC & 93.9 & 93.7 & 102.4 & 81.4 & 82.0 & 89.8 \\
        BayesSeg       & cAE & 99.0 & 99.9 & 122.0 & 82.6 & 85.4 & 104.7 \\
        BayesSegMinDur & cAE & 94.0 & 93.7 & 95.0 & 82.4 & 83.3 & 81.1 \\
        \bottomrule
    \end{tabularx}
    \label{tbl:bucktsong_si_wer}
\end{table*}

\begin{table}[tbp]
    \mytable
    \caption{Average speaker-independent cluster (clust.), speaker (spk.), and gender (gndr) purity for BayesSegMinDur on the three datasets.}
    \begin{tabularx}{\linewidth}{@{}c@{\ \ }CCcCCcCCc@{}}
        \toprule
        & \multicolumn{3}{c}{English1 (\%)} & \multicolumn{3}{c}{English2 (\%)} & \multicolumn{3}{c}{Xitsonga (\%)} \\
        \cmidrule(r){2-4} \cmidrule(lr){5-7} \cmidrule(l){8-10}
        Embeds. & Clust. & Spk. & Gndr & Clust. & Spk. & Gndr & Clust. & Spk. & Gndr \\
        \midrule
        MFCC & 30.3 & 56.7 & 86.8 & 29.9 & 55.9 & 87.6 & 24.5 & 43.1 & 87.1 \\
        cAE & 31.5 & 37.9 & 77.0 & 30.0 & 35.7 & 73.8 & 33.1 & 29.3 & 76.6 \\
        \bottomrule
    \end{tabularx}
    \label{tbl:bucktsong_si_purity}
\end{table}

\subsubsection*{Qualitative analysis and summary}

Qualitative analysis involved concatenating and listening to the audio from the tokens in some of the biggest clusters of the best speaker-dependent and -independent models.
Apart from the trends mentioned already, others also became immediately apparent.
Despite the low average cluster purity ranging from 30\% to 60\% in the analyses above, we found that most of the clusters are acoustically very pure: often tokens correspond to the same syllable or partial word, but occur within different ground truth words. For example, a cluster with the word `day' had the corresponding portions from `daycare' and `Tuesday'.
These are marked as errors for cluster purity and WER calculations.
In the next section, we use NED as metric, which does not penalize such partial word matches.
The biggest clusters often correspond to filler-words.
As an example, speaker S38 from English1 had several clusters corresponding to `yeah' and `you know'.  But the BayesSegMinDur-MFCC model applied to S38 also discovered pure clusters corresponding to `different', `people' and `five'.
For the speaker-independent BayesSegMinDur-cAE system, the biggest clusters consisted of instances of `um', `uh', `oh', `so' and `yeah'.

In summary\edit{, the high error rates reported above indicate that significant effort is still required in order to achieve reasonable performance with such zero-resource methods. A comparison of Tables~\ref{tbl:bucktsong_sd_wer} and~\ref{tbl:bucktsong_si_wer} shows that dealing with multiple speakers is particularly challenging---recent zero-resource work has started to investigate this aspect specifically~\cite{zeghidour+etal_interspeech16}.
Nevertheless, the above analysis allowed us to compare and draw conclusions regarding the different variants of our approach. Specifically, }
although under-segmentation occurs in the BayesSeg and BayesSegMinDur models,
these models yield more accurate clusters and thereby improve overall purity and WER.
In most cases, cAE embeddings either yield similar or improved performance compared to MFCCs.
In particular in the speaker-independent case, cAE-based models discover clusters that are more speaker- and gender-independent.
This illustrates the benefit of incorporating weak top-down supervision for unsupervised representation learning within a zero-resource system.

\subsection{Results: Comparison to other systems}
\label{sec:bucktsong_comparison}

We now compare our approach to others using
the evaluation framework provided as part of the ZRS challenge~\cite{versteegh+etal_interspeech15}.
We compare our approach to three systems:

\textbf{ZRSBaselineUTD} is the UTD system used as official baseline in the challenge~\cite{versteegh+etal_interspeech15}  (see Section~\ref{sec:background_utd}).

\textbf{UTDGraphCC} is the best UTD system of~\cite{lyzinski+etal_interspeech15}, employing a connected component graph clustering algorithm to group discovered segments (also Section~\ref{sec:background_utd}).

\textbf{SyllableSegOsc$^{\text{+}}$} uses oscillator-based syllabification followed by speaker-dependent clustering and word discovery~\cite{rasanen+etal_interspeech15} (Section~\ref{sec:background_full_coverage}).
We add the superscript + since, after publication of~\cite{rasanen+etal_interspeech15}, R{\"a}s{\"a}nen et al.\ further refined their syllable boundary detection method~\cite{rasanen+etal_submission16}. 
We use this updated version for presegmentation in our system. The authors of~\cite{rasanen+etal_interspeech15} kindly regenerated their full ZRS results for comparison here.
The original results are included in \iftoggle{csl}{\ref{appen:results_zrs}}{Appendix~\ref{appen:results_zrs}}.

For our approach, we focus on systems that performed best on English1 in the previous section: for the speaker-dependent setting we use the MFCC-based BayesSegMinDur system, while for the speaker-independent setting we use the cAE-based BayesSegMinDur model.
The performance of all our system variants using all of the ZRS metrics are given in \iftoggle{csl}{\ref{appen:results_zrs}}{Appendix~\ref{appen:results_zrs}}.

Figure~\ref{fig:ned_zrs} shows the NED scores of the different systems on English2 and Xitsonga.
ZRSBaselineUTD yields the best NED on both languages, with UTDGraphCC also performing well.
UTD systems like these explicitly aim to discover high-precision clusters of isolated segments, but do not cover all the data.
They are therefore tailored to NED, which only evaluates the patterns discovered by the method and does not evaluate recall on the rest of the data.
In contrast, SyllableSegOsc$^{\text{+}}$ and our own systems perform full-coverage segmentation.
Of these, our systems achieve better NED than SyllableSegOsc$^{\text{+}}$ on both languages, indicating that the discovered clusters in our approach are more consistent.
Even when running our system in a speaker-independent setting (BayesSegMinDur-cAE in the figure), our approach outperforms the speaker-dependent SyllableSegOsc$^{\text{+}}$.

Figures~\ref{fig:english_zrs} and~\ref{fig:xitsonga_zrs} show the token, type and boundary $F$-scores on the two languages.
\edit{For comparison, word token $F$-scores of less than $4\%$ were achieved at the 2012 JHU CSLP workshop, although a different dataset was used~\cite{jansen+etal_icassp13}.}
Apart from word type $F$-score on Xitsonga, our models outperform all other approaches \edit{in the direct comparison here}.
The UTD systems struggle on these metrics since the $F$-scores are based on precision and recall over the entire input.
The full-coverage SyllableSegOsc$^{\text{+}}$ is therefore our strongest competitor in most cases.
The prediction of word candidates from reoccurring cluster sequences in SyllableSegOsc$^{\text{+}}$ is done greedily and bottom-up, without regard to other word mappings in an utterance.
In contrast, BayesSegMinDur samples word boundaries and cluster assignments together by taking a whole utterance into account; it imposes a consistent top-down segmentation, while simultaneously adhering to bottom-up syllable boundary detection and minimum duration constraints.
The result is a more accurate segmentation of the data.
Note that in BayesSeg it is easy to incorporate additional bottom-up constraints (such as a minimum duration) and these are considered jointly with segmentation.
In contrast, such a minimum duration constraint would require additional heuristics in the pure bottom-up approach of~\cite{rasanen+etal_interspeech15}.

\begin{figure}[!t]
    \centering
    \iftoggle{csl}{\includegraphics[width=0.8\linewidth]{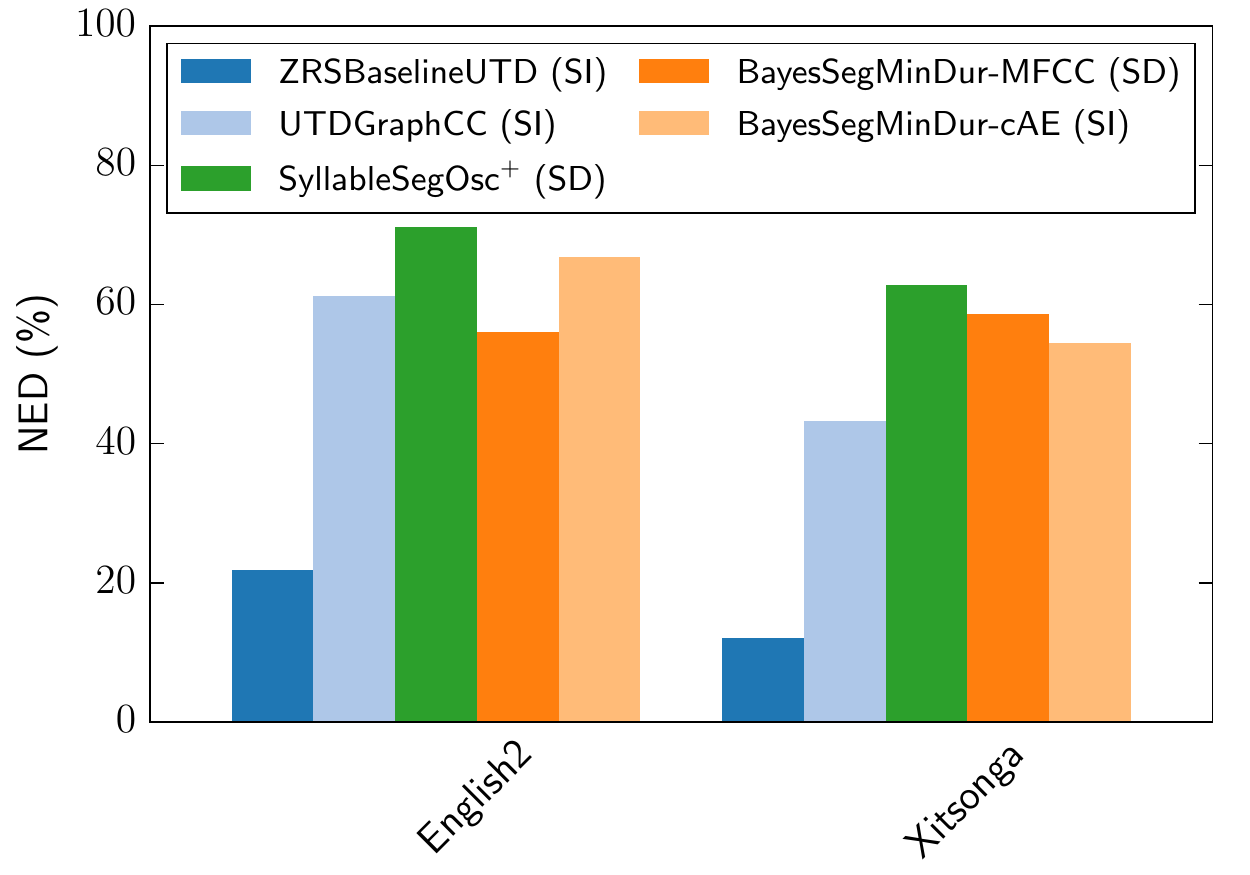}}{\includegraphics[width=0.95\linewidth]{ned_zrs}}
    \caption{Normalized edit distance (NED) on English2 and Xitsonga. Lower NED is better. Scores are only computed on the analyzed portion of data (so the lower-coverage UTD systems have an advantage). SD/SI indicates that a system is speaker-dependent/speaker-independent.}
    \label{fig:ned_zrs}
\end{figure}

\iftoggle{csl}{
    \begin{figure}[!p]
}{
    \begin{figure}[!t]
}

    \centering
    \iftoggle{csl}{\includegraphics[width=0.8\linewidth]{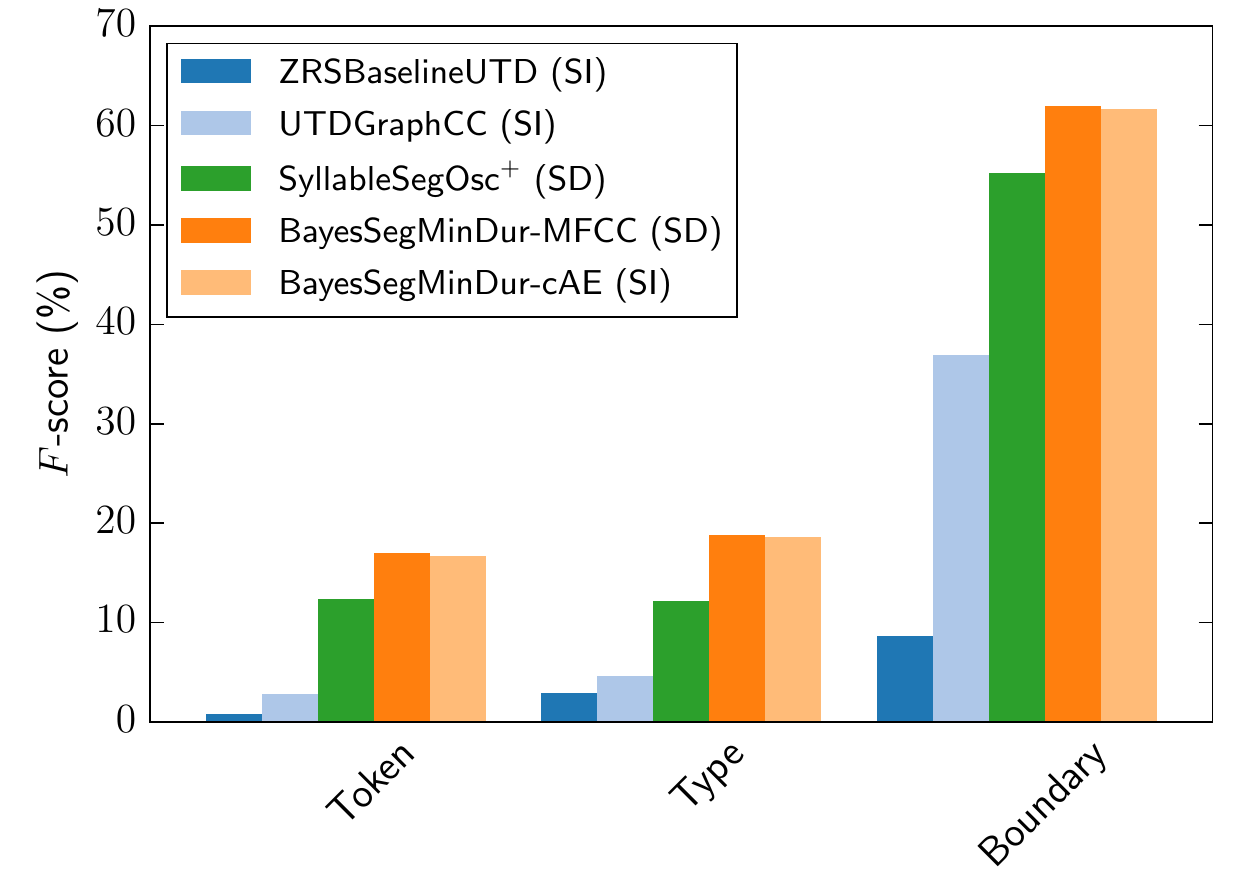}}{\includegraphics[width=0.95\linewidth]{english_zrs}}
    \caption{Word token, type and boundary $F$-scores on English2.
    SD/SI indicates that a system is speaker-dependent/speaker-independent.
    The word boundary detection tolerance is 30~ms or 50\% of a phoneme.
    }
    \label{fig:english_zrs}
\end{figure}

The results in Figures~\ref{fig:english_zrs} and~\ref{fig:xitsonga_zrs} also indicate that our speaker-independent system performs on par with the speaker-dependent system on these metrics; despite less accurate clusters (in terms of purity, WER and NED), the speaker-independent models still yields an accurate segmentation of the data, outperforming both speaker-independent UTD baselines and the speaker-dependent SyllableSegOsc$^{\text{+}}$.

We conclude that by hypothesizing word boundaries consistently over an utterance rather than taking these decisions in isolation,
our approach yields more accurate clusters (NED) that correspond better to true words (word type $F$-score) than the full-coverage syllable-based approach of~\cite{rasanen+etal_interspeech15}.
It also segments the data more accurately (word token and boundary $F$-scores), even when applying the model to data from multiple speakers.
However, despite the benefits of our model, the algorithm of~\cite{rasanen+etal_interspeech15} is much simpler in terms of computational complexity and implementation.
Compared to UTD systems which aim to find high-quality reoccurring patterns but do not cover all the data, the items in our clusters have a poorer match to each other (NED), but correspond better to true words on the English data (word type $F$-score). On both languages, our full-coverage method also segments the data better into word-like units (word boundary and token $F$-scores) than the UTD systems.

\section{Conclusion}

We presented a segmental Bayesian model which segments and clusters conversational speech audio---\edit{a first attempt to evaluate a full-coverage zero-resource system on multi-speaker large-vocabulary data.}
The system limits word boundary positions by using a bottom-up presegmentation method to detect syllable-like units, and relies on a segmental approach where word segments are represented as fixed-dimensional acoustic word embeddings.

Our speaker-dependent system achieves WERs of around 84\% on English and 76\% on Xitsonga data, outperforming a purely bottom-up method that treats each syllable as a word candidate.
Despite much worse speaker-independent performance, here we achieve improvements by incorporating frame-level features from an autoencoder-like neural network trained using weak top-down constraints.
This results in clusters that are purer and less speaker- and gender-specific than when using MFCCs\edit{, showing for the first time the benefit of unsupervised representation learning within
a complete zero-resource system.}

\iftoggle{csl}{
    \begin{figure}[!p]
}{
    \begin{figure}[!t]
}
    \centering
    \iftoggle{csl}{\includegraphics[width=0.8\linewidth]{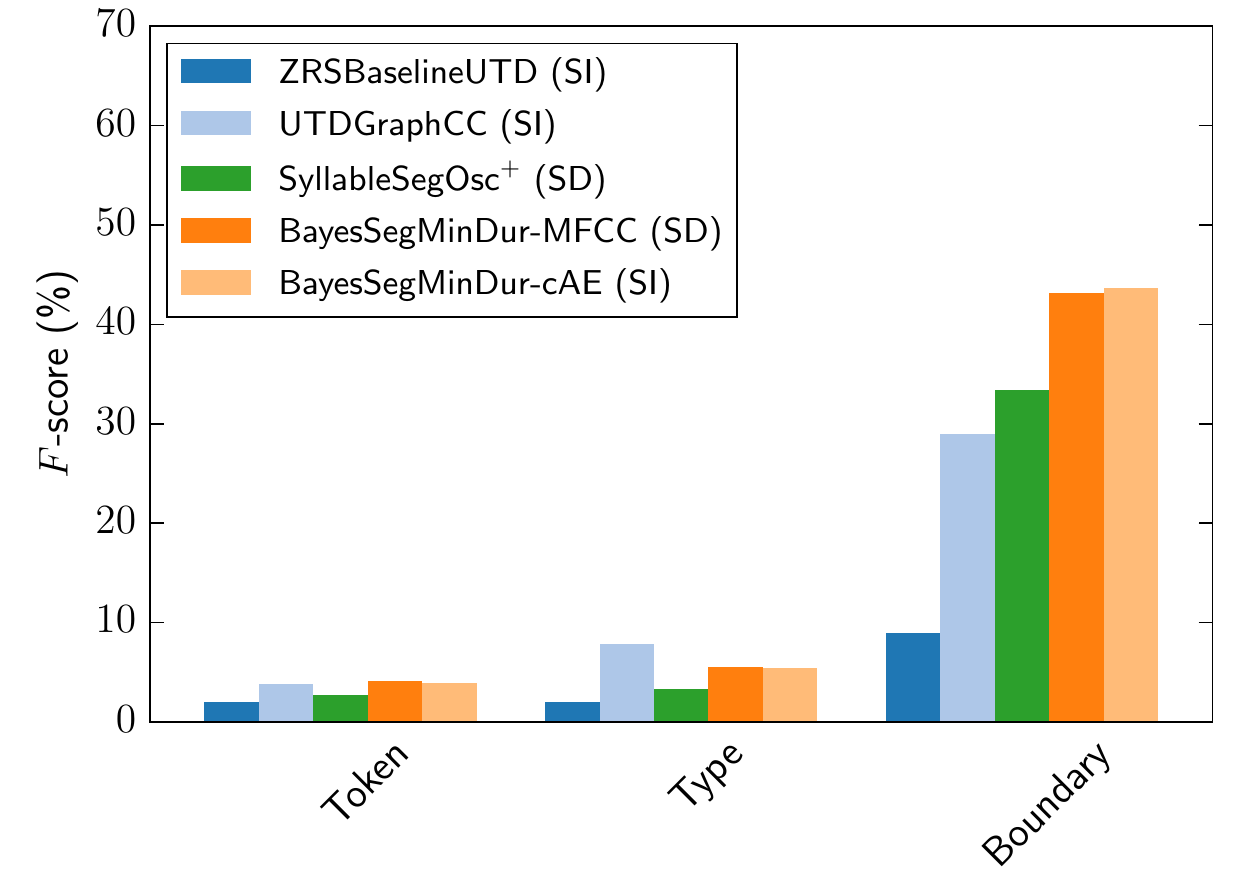}}{\includegraphics[width=0.95\linewidth]{xitsonga_zrs}}    
    \caption{Word token, type and boundary $F$-scores on Xitsonga.
    SD/SI indicates that a system is speaker-dependent/speaker-independent.
    The word boundary detection tolerance is 30~ms or 50\% of a phoneme.}
    \label{fig:xitsonga_zrs}
\end{figure}

We compared our approach to state-of-the-art baselines on both languages.
We found that, although the isolated patterns discovered by UTD are more consistent, the clusters of our full-coverage approach are better matched to true words, measured in terms of word token, type and boundary $F$-scores.
We also found that by proposing a consistent segmentation and clustering over whole utterances, our approach outperforms a purely bottom-up syllable-based full-coverage system on these metrics.

\edit{The high WERs reported in this study show that there is still much work to be done in the area of zero-resource speech processing. Nevertheless, previous work shows that high-error rate unsupervised systems can still be useful in downstream tasks.
The analysis presented here also provides useful baselines and guidance for future work.
In particular, we show the benefits of performing consistent top-down segmentation while adhering to bottom-up constraints, as well as incorporating unsupervised representation learning. Our own f}uture work will consider better acoustic word embedding approaches, improving the recall of the syllabic presegmentation method, and improving the overall efficiency of the model.

\section*{Acknowledgements}

We would like to thank Okko R{\"a}s{\"a}nen and Shreyas Seshadri for providing the code for their syllable boundary detection algorithm and for regenerating their ZRS results.
We also thank Roland Thiolli{\`e}re and Maarten Versteegh for providing us the alignments used in the ZRS challenge.
HK was funded by a Commonwealth Scholarship. This work was supported in part by a James S.\ McDonnell Foundation Scholar Award to SG.

\small
\bibliography{bucktsong2016}
\normalsize

\appendix
\section*{Appendices}
\titleformat*{\section}{\normalsize\bf}
\section{Posterior predictive of spherical Gaussian}
\label{appen:posterior_predictive}

\edit{
Because of the conjugate priors with known spherical covariance matrices, the probability density function (PDF) of the multivariate posterior predictive $p(\vec{x}_i | \mathcal{X}_{k \backslash i}; \vec{\beta})$ in~\eqref{eq:bucktsong_collapsed1}  is itself a spherical covariance Gaussian.
This PDF decomposes into the product of univariate PDFs; for a single dimension $x_i$ of vector $\vec{x}_i$, the univariate PDF is given by
\begin{equation}
    p(x_i|\mathcal{X}_{k \backslash i})
    = \mathcal{N} (x_i|\mu_{N_{k \backslash i}}, \sigma_{N_{k \backslash i}}^2 + \sigma^2) \label{eq:univariate_post_predict}
\end{equation}
where
\begin{equation}
    \sigma_{N_{k \backslash i}}^2 = \frac{\sigma^2\sigma_0^2}{N_{k \backslash i}\sigma_0^2 + \sigma^2} \text{\ \ ,\ \ }
    \mu_{N_{k \backslash i}} = \sigma_{N_{k \backslash i}}^2 \left( \frac{\mu_0}{\sigma_0^2} + \frac{N_{k \backslash i}\overline{x}_{k \backslash i}}{\sigma^2} \right)
\end{equation}
and $\overline{x}_{k \backslash i}$ is component $k$'s sample mean for this dimension~\cite{murphy_bayesgauss07}.
}

\section{Forward filtering backward sampling for word segmentation}
\label{appen:forward_backward}

To sample the new set of embeddings in line~\ref{alg_line:bucktsong_sample_bounds} of Algorithm~\ref{alg:bucktsong_gibbs_wordseg}, the forward filtering backward sampling dynamic programming algorithm is used~\cite{scott_jasa02}.
Forward variable $\alpha[t]$ is defined as the density of the frame sequence $\vec{y}_{1:t}$, with the last frame the end of a word: $\alpha[t] \defeq p(\vec{y}_{1:t} | {h^-})$. {The embeddings and component assignments for all words not in the current utterance $\vec{s}_i$, and the hyperparameters of the GMM, are denoted as $h^- = (\mathcal{X}_{\backslash s}, \vec{z}_{\backslash s}; a, \vec{\beta})$.}
The forward variables can be recursively calculated as~\cite{mochihashi+etal_acl09}:
\begin{align}
    \alpha[t]
    = \sum_{j = 1}^t p(\vec{y}_{{t - j + 1}:t} | h^-) \alpha[t - j] \label{eq:bucktsong_forward}
\end{align}
{starting with $\alpha[0] = 1$ and calculating~\eqref{eq:bucktsong_forward}
for $1 \leq t \leq M - 1$.}
The $p(\vec{y}_{{t - j + 1}:t} | h^-)$ term in~\eqref{eq:bucktsong_forward} is the value of a joint probability density function (PDF) over acoustic frames $\vec{y}_{{t - j + 1}:t}$.
In analogy to a frame-based supervised model where this term would be calculated as the product of the PDF values of a GMM for all the frames involved, we define this term as
\begin{equation}
    p(\vec{y}_{{t - j + 1}:t} | h^-) \defeq \left[p \left(\vec{x}' | h^- \right) \right]^j \label{eq:bucktsong_prob_segment}
\end{equation}
where $\vec{x}' = f_e(\vec{y}_{{t - j + 1}:t})$ is the acoustic word embedding calculated on the segment.
Thus, as in the frame-based supervised case, each frame is assigned a PDF score; but in this case, all $j$ frames in the segment are assigned the PDF value of the whole segment under the current acoustic model.
The required marginal term in~\eqref{eq:bucktsong_prob_segment} can be calculated as:
\begin{align}
    p(\vec{x}' | h^-)
    &= \sum_{k = 1}^{K} P(z_h = k | \vec{z}_{\backslash h}; a)  p(\vec{x}'| \mathcal{X}_{k\backslash h}; \vec{\beta})
    \label{eq:bucktsong_likelihood_fbgmm}
\end{align}
with the two terms in the summation calculated in the same way as those in~\eqref{eq:bucktsong_collapsed1}.

Once all $\alpha$'s have been calculated, a segmentation can be sampled backwards. 
Starting from the final positition $t = M$, we sample the preceding word boundary position using~\cite{mochihashi+etal_acl09}:
\begin{equation}
    P(q_t = j | \vec{y}_{1:t}, h^-) \propto p(\vec{y}_{{t - j + 1}:t} | h^-) \alpha[t - j]
    \label{eq:bucktsong_backward}
\end{equation}
Variable $q_t$ is the number of frames that we need to move backwards from position $t$ to find the preceding word boundary.
We calculate~\eqref{eq:bucktsong_backward} for $1 \leq j \leq t$ and sample while $t - j \geq 1$.

\section{Tables of complete results for all systems and metrics}
\label{appen:results_zrs}

In Section~\ref{sec:bucktsong_results_own}, several variants of our approach were considered. In Section~\ref{sec:bucktsong_comparison}, a subset of these were compared to other systems evaluated in the context of the Zero Resource Speech Challenge 2015 (ZRS)~\cite{versteegh+etal_interspeech15}, using a subset of the challenge metrics.
Tables~\ref{tbl:appen_zseval_english2} and~\ref{tbl:appen_zseval_tsonga} give the performance of all variants of our system on all the ZRS metrics on the English and Xitsonga data, respectively.

\begin{table*}[!p]
    \mytable
    \footnotesize
    \iftoggle{csl}{\begin{minipage}[c]{1.25\linewidth}}{}
    \caption{Performance of several systems on English2. All scores are given as percentages~(\%). The word boundary detection tolerance is 30~ms or 50\% of a phoneme.
    }
    \begin{tabularx}{\linewidth}{@{}l@{\ }CCCCCCCCCCCCCC@{}}
        \toprule
        & \multicolumn{2}{c}{NLP} & \multicolumn{3}{c}{Grouping} & \multicolumn{3}{c}{Word token} & \multicolumn{3}{c}{Word type} & \multicolumn{3}{c@{}}{Word boundary} \\
        \cmidrule{2-3} \cmidrule(l){4-6} \cmidrule(l){7-9} \cmidrule(l){10-12} \cmidrule(l){13-15}
        Model & NED & Cov. & Prec. & Rec. & $F$ & Prec. & Rec. & $F$ & Prec. & Rec. & $F$ & Prec. & Rec. & $F$  \\
        \midrule
        \multicolumn{15}{@{}l}{\textit{Systems from previous studies:}} \\
        ZRSTopline~\cite{versteegh+etal_interspeech15} & 0 & 100 & 99.5 & 100 & 99.7 & 68.2 & 60.8 & 64.3 & 50.3 & 56.2 & 53.1 & 88.4 & 86.7 & 87.5 \\
        ZRSBaselineUTD~\cite{versteegh+etal_interspeech15} & \textbf{21.9} & 16.3 & 21.4 & \textbf{84.6} & \textbf{33.3} & 5.5 & 0.4 & 0.8 & 6.2 & 1.9 & 2.9 & 44.1 & 4.7 & 8.6 \\
        UTDGraphCC~\cite{lyzinski+etal_interspeech15} & 61.2& 80.2& - & - & - & 2.4 & 3.5 & 2.8 & 3.1 & 9.2 & 4.6 & 35.4& 38.5& 36.9 \\
        SyllableSegOsc~\cite{rasanen+etal_interspeech15} & 70.8 & 42.4 & 13.4 & 15.7 & 14.2 & 22.6 & 6.1 & 9.6 & 14.1 & 12.9 & 13.5 & 75.7 & 33.7 & 46.7 \\
        SyllableSegOsc$^{\text{+}}$ & 71.1 & 100 & 10.2 & 16.3 & 12.6 & 14.3 & 10.9 & 12.4 & 8.4 & 22.1 & 12.2 & 61.1 & 50.1 & 55.2 \\
        \addlinespace
        \multicolumn{15}{@{}l}{\textit{Speaker-dependent, MFCC embeddings:}} \\
        SyllableBayesClust & 62.2 & 100 & 17.5 & 11.2 & 13.7 & 21.5 & 18.0 & 19.6 & 12.3 & 28.8 & 17.2 & 63.8 & \textbf{59.8} & 61.7 \\
        BayesSeg & 61.5 & 100 & 17.1 & 13.7 & 15.2 & 24.0 &  18.1 & \textbf{20.6} & 13.1 & \textbf{30.1} & 18.2 & 67.3 & 58.3 & \textbf{62.5} \\
        BayesSegMinDur & 56.0 & 100 & 22.7 & 29.6 & 25.5 & 26.6 & 12.5 & 17.0 & 14.0 & 28.6 & \textbf{18.8} & 80.7 & 50.4 & 62.0 \\
        \addlinespace
        \multicolumn{15}{@{}l}{\textit{Speaker-dependent, cAE embeddings:}} \\
        BayesSeg & 62.1 & 100 & 18.0 & 15.0 & 16.3 & 24.8 & 17.0 & 20.2 & 13.3 & 29.1 & 18.3 & 69.4 & 56.3 & 62.2 \\
        BayesSegMinDur & 57.2 & 100 & \textbf{23.7} & 26.3 & 24.9 & \textbf{27.6} & 11.9 & 16.6 & \textbf{14.2} & 26.7 & 18.5 & \textbf{83.1} & 49.0 & 61.6 \\
        \addlinespace
        \multicolumn{15}{@{}l}{\textit{Speaker-independent, MFCC embeddings:}} \\
        SyllableBayesClust & 73.0 & 100 & 9.2 & 5.1 & 6.5 & 21.5 & 18.0 & 19.6 & 12.3 & 28.8 & 17.2 & 63.8 & \textbf{59.8} & 61.7 \\
        BayesSeg & 73.2 & 100 & 9.1 & 5.9 & 7.2 & 23.6 & \textbf{18.2} & \textbf{20.6} & 12.8 & 29.6 & 17.9 & 66.5 & 58.8 & 62.4 \\
        BayesSegMinDur & 72.0 & 100 & 9.9 & 13.0 & 11.2 & 25.9 & 12.6 & 17.0 & 13.7 & 28.9 & 18.6 & 79.7 & 51.4 & 62.1 \\
        \addlinespace
        \multicolumn{15}{@{}l}{\textit{Speaker-independent, cAE embeddings:}} \\
        BayesSeg & 71.1 & 100 & 10.3 & 7.2 & 8.5 & 24.5 & 16.6 & 19.8 & 12.9 & 27.7 & 17.6 & 69.6 & 55.8 & 62.0 \\
        BayesSegMinDur & 66.9 & 100 & 11.9 & 14.0 & 12.8 & 26.9 & 12.2 & 16.7 & 14.1 & 27.5 & 18.6 & 81.7 & 49.6 & 61.7 \\
        \bottomrule
    \end{tabularx}
    \label{tbl:appen_zseval_english2}
    \iftoggle{csl}{\end{minipage}}{}
\end{table*}

\begin{table*}[!p]
    \mytable
    \footnotesize
    \iftoggle{csl}{\begin{minipage}[c]{1.25\linewidth}}{}
    \caption{Performance of several systems on Xitsonga. All scores are given as percentages~(\%). The word boundary detection tolerance is 30~ms or 50\% of a phoneme.
    }
    \begin{tabularx}{\linewidth}{@{}l@{\ }CCCCCCCCCCCCCC@{}}
        \toprule
        & \multicolumn{2}{c}{NLP} & \multicolumn{3}{c}{Grouping} & \multicolumn{3}{c}{Word token} & \multicolumn{3}{c}{Word type} & \multicolumn{3}{c@{}}{Word boundary} \\
        \cmidrule{2-3} \cmidrule(l){4-6} \cmidrule(l){7-9} \cmidrule(l){10-12} \cmidrule(l){13-15}
        Model & NED & Cov. & Prec. & Rec. & $F$ & Prec. & Rec. & $F$ & Prec. & Rec. & $F$ & Prec. & Rec. & $F$  \\
        \midrule
        \multicolumn{15}{@{}l}{\textit{Systems from previous studies:}} \\
        ZRSTopline~\cite{versteegh+etal_interspeech15} & 0 & 100 & 100 & 100 & 100 & 34.1 & 49.7 & 40.4 & 15.1 & 18.1 & 16.5 & 66.6 & 91.9 & 77.2 \\
        ZRSBaselineUTD~\cite{versteegh+etal_interspeech15} & \textbf{12.0} & 16.2 & \textbf{52.1} & \textbf{77.4} & \textbf{62.2} & 3.2 & 1.4 & 2.0 & 3.2 & 1.4 & 2.0 & 22.3 & 5.6 & 8.9\\
        UTDGraphCC~\cite{lyzinski+etal_interspeech15} & 43.2 & 89.4 & - & - & - & 2.2 & \textbf{12.6} & 3.8 & \textbf{4.9} & \textbf{18.8} & \textbf{7.8} & 18.8 & \textbf{64.0} & 29.0\\
        SyllableSegOsc~\cite{rasanen+etal_interspeech15} & 63.1 & 94.7 & 10.7 & 3.3 & 5.0 & 2.3 & 3.4 & 2.7 & 2.2 & 6.2 & 3.3 & 29.2 & 39.4 & 33.5 \\
        SyllableSegOsc$^{\text{+}}$ & 62.8 & 94.7 & 10.6 & 3.1 & 4.8 & 2.3 & 3.3 & 2.7 & 2.3 & 6.3 & 3.3 & 29.1 & 39.1 & 33.4 \\
        \addlinespace
        \multicolumn{15}{@{}l}{\textit{Speaker-dependent, MFCC embeddings:}} \\
        SyllableBayesClust & 57.7 & 100 & 13.0 & 2.5 & 4.2 & 3.8 & 6.8 & 4.9 & 2.5 & 6.6 & 3.6 & 31.4 & 52.3 & 39.2 \\
        BayesSeg & 56.5 & 100 & 12.7 & 4.1 & 6.2 & 4.1 & 6.2 & 4.9 & 2.9 & 7.8 & 4.2 & 34.5 & 49.0 & 40.5 \\
        BayesSegMinDur & 58.6 & 100 & 8.3 & 10.3 & 9.2 & \textbf{4.3} & 4.0 & 4.1 & 3.8 & 9.8 & 5.5 & 44.5 & 42.0 & 43.2  \\
        \addlinespace
        \multicolumn{15}{@{}l}{\textit{Speaker-dependent, cAE embeddings:}} \\
        BayesSeg & 52.6 & 100 & 16.0 & 5.0 & 7.6 & 4.1 & 5.7 & 4.8 & 3.1 & 8.1 & 4.5 & 36.0 & 47.5 & 41.0 \\
        BayesSegMinDur & 57.0 & 100 & 10.3 & 13.6 & 11.7 & 4.2 & 3.4 & 3.7 & 3.7 & 9.3 & 5.3 & \textbf{47.8} & 40.6 & \textbf{43.9} \\
        \addlinespace
        \multicolumn{15}{@{}l}{\textit{Speaker-independent, MFCC embeddings:}} \\
        SyllableBayesClust & 63.0 & 100 & 8.8 & 3.5 & 5.0 & 3.8 & 6.8 & 4.9 & 2.5 & 6.6 & 3.6 & 31.4 & 52.3 & 39.2 \\
        BayesSeg & 63.6 & 100 & 7.7 & 4.4 & 5.6 & 4.1 & 6.5 & \textbf{5.0} & 2.7  &7.4  &4.0 & 33.5 & 50.0 & 40.1  \\
        BayesSegMinDur & 64.8 & 100 & 4.8 & 8.1 & 6.0 & 3.9 & 3.9 & 3.9 & 3.5 & 9.2 & 5.0 & 42.4 & 42.5 & 42.4 \\
        \addlinespace
        \multicolumn{15}{@{}l}{\textit{Speaker-independent, cAE embeddings:}} \\
        BayesSeg & 55.4 & 100 & 12.6 & 12.8 & 12.7 & 4.2 & 5.3 & 4.7 & 3.1 & 8.1 & 4.5 & 37.6 & 46.2 & 41.5 \\
        BayesSegMinDur & 54.5 & 100 & 9.4 & 21.1 & 13.0 & 4.2 & 3.6 & 3.9 & 3.8 & 9.5 & 5.4 & 46.5 & 41.2 & 43.7 \\
        \bottomrule
    \end{tabularx}
    \label{tbl:appen_zseval_tsonga}
    \iftoggle{csl}{\end{minipage}}{}
\end{table*}

\end{document}